\documentclass[11pt]{article}

\usepackage[preprint]{acl}

\usepackage{times}
\usepackage{latexsym}
\usepackage[T1]{fontenc}
\usepackage[utf8]{inputenc}
\usepackage{microtype}
\usepackage{inconsolata}
\usepackage{graphicx}

\providecommand{\Description}[1]{}

\usepackage{amsthm}
\usepackage{amsmath}
\usepackage{amssymb}
\usepackage{subcaption}
\usepackage{paralist}
\usepackage{cases}
\usepackage{booktabs}
\usepackage{threeparttable}
\usepackage{epstopdf}
\usepackage{upgreek}
\usepackage{endnotes}
\usepackage{etoolbox}
\usepackage{algpseudocode}
\usepackage{xspace}
\usepackage{array}
\usepackage{enumitem}
\usepackage{balance}
\usepackage{multirow}
\usepackage{xcolor}
\usepackage{tcolorbox}
\tcbuselibrary{skins,breakable}
\usepackage{listings}
\usepackage{wrapfig}
\usepackage{tabularx}
\usepackage{float}
\usepackage{flushend}
\usepackage[ruled,linesnumbered]{algorithm2e}
\usepackage{stfloats}
\usepackage{afterpage}
\usepackage{multicol}

\theoremstyle{definition}
\newtheorem{definition}{Definition}

\definecolor{deepgreen}{RGB}{34,139,34}

\newcommand{\ie}{\emph{i.e.,}\xspace}
\newcommand{\eg}{\emph{e.g.,}\xspace}

\title{TrafficClaw: A Generalizable LLM Agent in the Unified Physical Environment for Urban Traffic Control}

\author{Siqi Lai$^{1}$ \quad Pan Zhang$^{1}$ \quad Yuping Zhou$^{1}$ \quad Jindong Han$^{2}$ \quad Yansong Ning$^{1}$ \quad Hao Liu$^{1}$ \\
  \texttt{slai125@connect.hkust-gz.edu.cn} \\[0.5em]
  $^{1}$The Hong Kong University of Science and Technology (Guangzhou) \\
  Guangzhou, Guangdong, China \\
  $^{2}$Shandong University, Jinan, Shandong, China}

\begin{document}
\maketitle

\begin{abstract}
  Large language model (LLM) agents have shown strong capabilities in long-horizon reasoning, tool use, and decision-making in digital environments, yet extending them to physically grounded systems remains challenging. Unlike web, code, or game environments, where objectives are often weakly coupled, physical systems evolve through tightly coupled dynamics in which local interventions propagate across interacting subsystems over time. Urban traffic control exemplifies this challenge, as traffic signals, freeways, public transit, and taxi systems continuously interact through shared spatial infrastructure and temporal mobility demand. Existing optimization, reinforcement learning (RL), and LLM-based approaches are largely designed for isolated subsystems, limiting coordinated reasoning and system-level optimization. We propose \textit{TrafficClaw}, a LLM-based generalizable traffic control agent for physical urban systems. TrafficClaw operates within a unified traffic environment that exposes coupled urban dynamics and feedback, performs executable spatiotemporal reasoning with persistent memory for long-horizon adaptation, and leverages multi-stage agentic RL for coordinated system-level optimization. Experiments across three metropolitan regions and six traffic-control tasks demonstrate strong generalization, robustness, and cross-subsystem coordination. Our project is available at \url{https://github.com/usail-hkust/TrafficClaw}.
\end{abstract}


\section{Introduction}

Large language model (LLM)-based systems have recently emerged as a promising paradigm for general-purpose agents, demonstrating strong capabilities in long-horizon planning, tool use, and reasoning~\cite{hu2025agentgen, xi2025agentgym, fang2025towards, pang2025browsemaster}. However, most existing LLM agents are developed in separate digital environments such as search engines~\cite{li2025webthinker}, games~\cite{dong2024villageragent}, and computer-use systems~\cite{openclaw2025}, where tasks remain modular, actions are mediated through predefined interfaces, and feedback is largely localized to the current environment state. Such environments provide limited support for modeling physical-world dynamics, where local interventions can propagate across interacting subsystems over time. Extending LLM agents from digital task execution to physical-world decision-making therefore requires environments that explicitly expose coupled state transitions, temporal feedback, and cross-subsystem interactions.

\begin{figure}[t]
    \centering
    \includegraphics[width=\columnwidth]{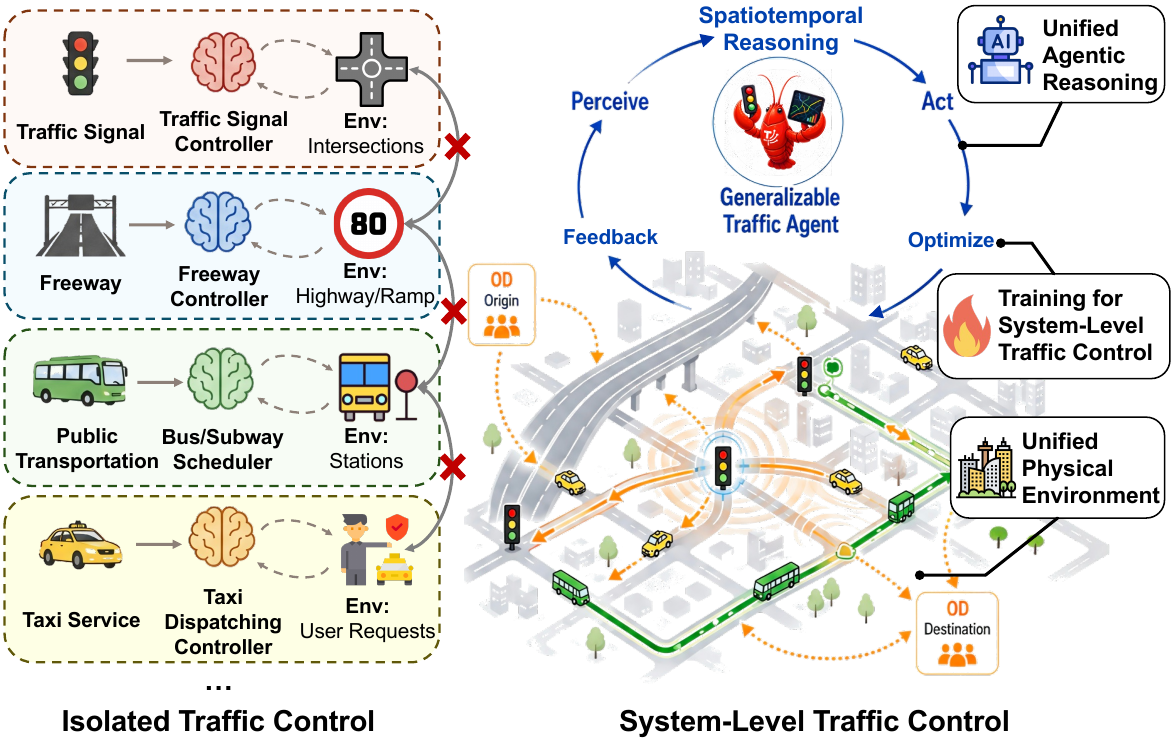}
    \caption{Isolated vs. unified traffic control.}
    \label{fig:motivation}
    \vspace{-10pt}
\end{figure}

Urban traffic exemplifies this challenge in physical environments. Heterogeneous subsystems (\eg traffic signals, freeways, public transit, taxi services) interact under shared spatial infrastructure and temporal mobility demand constraints. Existing traffic control methods, including optimization-based~\cite{koonce2008traffic,papageorgiou1991alinea}, RL-based~\cite{farazi2021deep,wang2022robust,jin2019coride,li2017reinforcement,belletti2017expert}, and recent LLM agents~\cite{lai2025llmlight,yuan2025collmlight,han2025large,feng2025agentmove}, are largely developed for isolated subsystems. As shown in Figure~\ref{fig:motivation}, such isolated formulations fail to capture coupled physical dynamics across subsystems, limiting coordinated reasoning and preventing system-level decision-making. These limitations motivate a unified agentic framework that embeds traffic control within a shared physical environment for cross-subsystem reasoning and system-level optimization.

Bringing LLM agents into urban traffic control raises three challenges. First, isolated subsystem environments cannot capture coupled urban mobility dynamics. Generalizable traffic control therefore requires a unified physical environment where heterogeneous subsystems interact through shared spatial infrastructure and temporal mobility demand constraints. Second, subsystems exhibit distinct spatial structures and temporal dynamics. Agents need to reason coherently across heterogeneous subsystems. Third, unified optimization remains difficult because different subsystems involve distinct objectives, reward scales, and potentially conflicting goals, where local improvements may degrade global performance. Effective traffic control therefore requires coordinated system-level optimization across heterogeneous subsystems.

To address these challenges, we propose \textit{TrafficClaw}, a LLM-based generalizable traffic control agent for physical urban systems. TrafficClaw operates within a unified physical traffic environment that integrates heterogeneous subsystems through shared spatial infrastructure and temporal mobility demand dynamics. The environment models coupled urban traffic dynamics, allowing local interventions to propagate across subsystems and induce network-wide feedback. This environment provides a shared interaction space for long-horizon reasoning, cross-subsystem coordination, and system-level optimization.

Within this unified environment, we develop a unified LLM agent for traffic control. Instead of using predefined pipelines for each subsystem, the agent performs unified analysis across spatial, temporal, and cross-subsystem dimensions through flexible code-based analytical modeling. This process generates structured diagnostics of how congestion, demand, and interventions propagate throughout the network, enabling coherent reasoning across heterogeneous subsystems. Meanwhile, the spatiotemporal memory accumulates reusable procedural knowledge, including congestion patterns, coordination strategies, and failure modes, across episodes. By integrating spatiotemporal reasoning with accumulated memory, TrafficClaw continuously self-improves its decision-making and adapts to diverse traffic dynamics.

To effectively translate analytical reasoning into actionable policies, we adopt a multi-stage training strategy. The agent is initialized using diverse multi-task trajectories spanning spatial regions, temporal patterns, and task configurations, bootstrapping generalizable reasoning across traffic subsystems. It is then optimized via agentic RL under system-level objectives that jointly model traffic efficiency and coordination quality. This interaction-driven training enables unified optimization across heterogeneous and potentially conflicting objectives, yielding system-aware control policies.

Our contributions are threefold. (1) We introduce an LLM agent that operates within a unified urban traffic environment, integrating heterogeneous subsystems into a shared physical system. To our knowledge, this is the first framework enabling generalizable traffic control with LLM agents in a unified environment. (2) We propose a traffic control agent with executable spatiotemporal reasoning, persistent memory mechanism, and system-level agentic RL, enabling unified diagnostics, long-horizon adaptation, and coordinated decision-making across coupled and heterogeneous subsystems. (3) Through extensive experiments on three metropolitan regions and six traffic-control tasks, we demonstrate that TrafficClaw achieves strong generalization and robustness across traffic subsystems and task configurations.

\section{Preliminary}

\begin{definition}
\textbf{Agentic Traffic Control}. 
We formulate traffic control as a partially observable Markov decision process (POMDP) defined by a tuple $\langle S, O, A, \mathcal{T}, \mathcal{F}, R \rangle$:
\begin{itemize}[leftmargin=0.5cm,itemsep=0pt,topsep=0.2em,parsep=0pt]
    \item \textbf{State}: $S$ is the traffic system state space, and $s_t \in S$ denotes the traffic state at time $t$ capturing spatiotemporal conditions of the controlled traffic infrastructure;
    \item \textbf{Observation}: $O$ is the observation space, and $o_t \in O$ denotes the observation at time $t$ obtained from sensors or monitoring systems, where $o_t \subseteq s_t$ is a subset of the state;
    \item \textbf{Action}: $A$ is the control action space, and $a_t \in A$ denotes the control action applied at time $t$ (\eg signal phase change);
    \item \textbf{Task}: $\tau \in \mathcal{T}$ is a traffic control task, specifying an operational goal such as flow optimization in traffic signal control;
    \item \textbf{Transition function}: $\mathcal{F}: S \times A \rightarrow S$ models the traffic dynamics, defining how the state evolves in response to control actions, capturing physical traffic flow dynamics and congestion propagation under the applied control;
    \item \textbf{Reward function}: $R(s_t, a_t) = f_{\text{reward}}(s_t, a_t)$ evaluates the immediate performance of executing action $a_t$ under state $s_t$.
\end{itemize}
Given a traffic control POMDP $\langle S, O, A, \mathcal{F}, R \rangle$ and a user query specifying a task $\tau \in \mathcal{T}$, agentic traffic control aims to perform automatic reasoning and decision-making to meet the user's request.
The agent observes historical traffic observations $\mathcal{O}_{1:t} = \{o_0, \ldots, o_t\}$ and past actions $\mathcal{A}_{1:t} = \{a_0, \ldots, a_{t-1}\}$. 
At each decision step $t$, the agent produces a sequence of control actions:
\begin{equation}
\mathcal{A}_{t:t+H} = \pi(\tau, \mathcal{O}_{1:t}, \mathcal{A}_{1:t}),
\end{equation}
where $\pi(\cdot)$ is the policy function that maps the input to the control action sequence. The goal is to optimize task-specific performance under physical and operational constraints.
\end{definition}


\section{Unified Traffic Environment}\label{sec:environment_construction}

\subsection{Urban Traffic System Modeling}

Urban traffic consists of heterogeneous subsystems coupled through spatial infrastructure and mobility demand constraints. We construct a unified traffic environment that integrates subsystems into a shared interactive framework for closed-loop perception, analysis, action, and optimization.

\begin{definition}
\textbf{Traffic Subsystem}. 
A traffic subsystem refers to a basic component of urban transportation that can be controlled to optimize traffic efficiency. Formally, subsystem $n$ is defined by its action space $A^{(n)}$ and action $a_t^{(n)}\in A^{(n)}$ at time $t$.
\end{definition}

\begin{definition}
\textbf{Unified Traffic Environment}.
A unified traffic environment integrates $N$ heterogeneous traffic subsystems into a single shared dynamical system operating over a common physical traffic network. It serves as the operational platform for the agentic traffic control process. Formally, it is defined by a tuple $\langle N, S^{env}, S^{I}, S^{M}, \{A^{(n)}, \Pi_{I}^{(n)}, \Pi_{M}^{(n)}\}_{n=1}^N, \mathcal{F} \rangle$:
\begin{itemize}[leftmargin=0.5cm,itemsep=0pt,topsep=0.2em,parsep=0pt]
    \item \textbf{Subsystems}: $N$ is traffic subsystem number;
    \item \textbf{Environment state}: $S^{env}$ is the global traffic state space over the entire traffic network;
    \item \textbf{Infrastructure}: $I$ denotes the common spatial infrastructure, including roads, lanes, stations, and zones. $S^{I}$ is its state space, with infrastructure state $s_t^{I}\in S^{I}$. $\Pi_{I}^{(n)}$ maps subsystem $n$ to the infrastructure it observes or controls;
    \item \textbf{Mobility}: $M$ denotes the time-indexed mobility domain. $S^{M}$ is its state space, with mobility state $s_t^{M} \in S^{M}$. $\Pi_{M}^{(n)}$ maps subsystem $n$ to the mobility it serves or reshapes;
    \item \textbf{Transition functions}: $\mathcal{F}^{env}$ is the unified transition function induced by local subsystem transitions over the shared infrastructure and demand states, formally defined as:
    \begin{align}
        &s_{t+1}^{I,(n)}
        = \mathcal{F}_{I}^{(n)}
           \bigl(s_t^{\text{env}}, \Pi_{I}^{(n)}s_t^{I}, a_t^{(n)}\bigr), \\[2pt]
        &s_{t+1}^{M,(n)}
        = \mathcal{F}_{M}^{(n)}
           \bigl(s_t^{\text{env}}, \Pi_{M}^{(n)}s_t^{M}, a_t^{(n)}\bigr), \\[2pt]
        &s_{t+1}^{\text{env}}
        = \mathcal{F}_{\text{env}}\!
           \bigl(s_t^{\text{env}}, s_{t}^{I},
                 s_{t}^{M},\{a_t^{(n)}\}_{n=1}^{N}\bigr),
    \end{align}
    where local transitions $\mathcal{F}_{I}^{(n)}$ and $\mathcal{F}_{M}^{(n)}$ describe how subsystem actions alter their associated infrastructure and mobility dynamics. If two subsystems operate on the same road, station, zone, or mobility demand, their local interventions induce a shared infrastructure-mobility state, which together reshapes the global environment.
\end{itemize} 
\end{definition}

\begin{definition}
\textbf{Urban Traffic Control}. 
Urban traffic control is defined as a POMDP tuple $\langle S^{env}, S^{I}, S^{M}, O, \{A^{(n)}\}_{n=1}^N, \mathcal{T}', \mathcal{F}^{env}, R \rangle$, where $\mathcal{T}' = \{\tau_k\}_{k=1}^K$ specifies $K (K \geq 1)$ control tasks from different subsystems. The goal is to optimize the local subsystem- and system-level performance under coupled network-wide dynamics through a runtime loop that continuously maps observation, analysis, action, and feedback into improved urban traffic efficiency.
\end{definition}

\subsection{Multi-Mode Mobility Demand Modeling}

Mobility demand links traveler behavior to subsystem dynamics, which specifies how travelers enter and interact with the shared physical traffic 
system across space and time \cite{xu2025mm,xu2026most}. We generate urban origin--destination (OD) demand from official government household travel surveys (\eg NYC Mobility Survey \cite{nyc_mobility_survey}), using empirical priors on trip volumes and mode preferences to ground demand in real-world behavior.

We model aggregate inter-region demand using a gravity-based formulation~\cite{anderson2011gravity, zhao2024origin, piao2025agentsociety, bougie2025citysim}. Let $i$ and $j$ denote origin and destination regions. The demand between them is modeled as the product of the activity levels of the two regions and is impeded by travel distance: 
\begin{equation}
D_{ij} \propto \frac{Q_i \cdot Q_j}{e_{ij}},
\end{equation}
where $Q_i$ and $Q_j$ denote regional activity intensities computed from population density and POI concentration, and $e_{ij}$ denotes the travel impedance (\ie free-flow travel time).

We assign the travel mode (\eg walk, vehicle, subway) of each trip from mobility survey statistics and construct the mode-specific OD demand as:
\begin{equation}
D_{ij}^{(m)} = D_{ij} \cdot p(m \mid c_{ij}),
\end{equation}
where $m \in \mathcal{M}$ denotes the travel mode and $p(m \mid c_{ij})$ denotes the probability of selecting mode $m$ conditioned on the trip purpose--distance category $c_{ij}$. These mode-specific demands initialize the mobility state $s_t^{\mathcal{M}}$, which is subsequently updated through coupled subsystem interactions. Consequently, mobility demand dynamically interferes with evolving traffic dynamics.

\section{Generalizable Traffic Control Agent}

\begin{figure*}[t]
\centering
\includegraphics[width=1.0\textwidth]{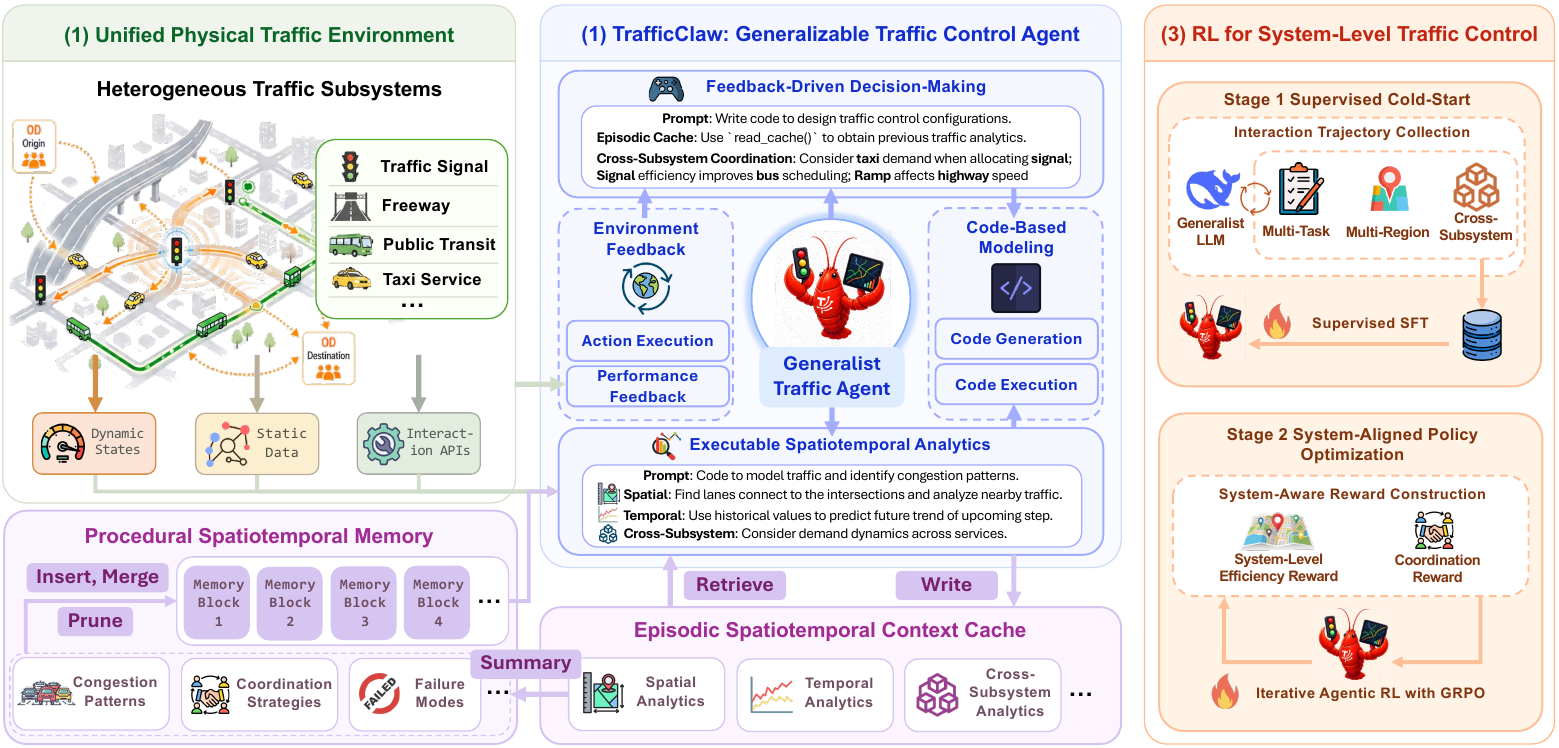}
\caption{The framework overview of TrafficClaw.}
\Description{Block diagram of the TrafficClaw framework: unified traffic environment, agent with spatiotemporal reasoning and memory, and multi-stage training connecting perception, analysis, action, and feedback.}
\label{fig:archi}
\vspace{-10pt} 
\end{figure*}

Figure~\ref{fig:archi} presents an overview of TrafficClaw for generalizable traffic control. 
(1) The agent performs executable spatiotemporal reasoning to generate structured diagnostics and feedback-grounded actions through closed-loop agent-environment interactions. 
(2) A spatiotemporal memory mechanism maintains cross-episode analytical context, accumulating procedural knowledge for self-improvement in long-horizon control.
(3) A multi-stage training strategy starts with supervised initialization and is followed by agentic RL under system-level objectives, enabling unified optimization across heterogeneous subsystems and aligning local decisions with system-level traffic efficiency.

\subsection{Agentic Spatiotemporal Reasoning}
\subsubsection{Executable Spatiotemporal Analytics}

To enable system-level traffic control, the agent needs to reason about (1) where inefficiencies emerge, (2) how they evolve over time, and (3) how local interventions affect system-level efficiency. Since predefined pipelines cannot adequately model heterogeneous dynamics and unseen scenarios, we formulate decision-making as an executable reasoning loop that generates and executes code for analytical modeling. This process produces structured diagnostics of coupled traffic dynamics through a unified analytical interface, enabling coherent cross-subsystem reasoning and coordinated decision-making.

Given $\mathcal{O}_{1:t}$ and $\mathcal{T}'$, the agent first generates and executes analytical code over the observed spatiotemporal traffic data. Concretely, the agent is instructed to generate executable code for analytical modeling, including:
(1) \emph{spatial modeling} $\Phi_{\mathrm{space}}(\cdot)$ captures network topology and spatial dependencies;
(2) \emph{temporal modeling} $\Phi_{\mathrm{time}}(\cdot)$ extracts dynamic patterns from traffic evolution; and
(3) \emph{cross-subsystem modeling} $\Phi_{\mathrm{cross}}(\cdot)$ characterizes how local interventions propagate across subsystems.
Rather than relying on a fixed pipeline, the agent implements the required analytics based on the task objectives and observed dynamics:
\begin{equation}
    Y_t^{\mathrm{ana}} = \Phi_{\mathrm{ana}}(\mathcal{O}_{1:t}, \mathcal{T}'),
\end{equation}
where $\Phi_{\mathrm{ana}}(\cdot)$ denotes a composition of analytical modeling processes (\eg $\Phi_{\mathrm{ana}} = \Phi_{\mathrm{space}} \circ \Phi_{\mathrm{time}} \circ \Phi_{\mathrm{cross}}$), and $Y_t^{\mathrm{ana}}$ is a structured diagnostic that summarizes the spatiotemporal state of the system and serves as the basis for downstream decision-making. The detailed prompt and analytical modeling processes are in Appendix~\ref{app:agent-prompt}.

\subsubsection{Feedback-Driven Decision-Making}

Based on the analytical outputs, the agent generates control actions grounded in inferred spatiotemporal traffic dynamics. Formally, given $Y_t^{\mathrm{ana}}$ and task set $\mathcal{T}'$, the actions at the next step are produced as:
\begin{equation}
\mathcal{A}_t = \Phi_{\text{act}}(Y_t^{\mathrm{ana}}, \mathcal{O}_{1:t}, \mathcal{T}'),
\end{equation}
where $\Phi_{\text{act}}(\cdot)$ denotes the executable code of the control policy generated by the agent. TrafficClaw evaluates them through a feedback-driven rollout process within the unified environment:
\begin{align}
    &\hat{s}_{t}^{\text{env}} = \mathcal{F}^{\text{env}}\!\bigl(s_{t-1}^{\text{env}}, \mathcal{A}_{t}\bigr), \\[4pt]
    &r_{t} =  \sum_{\tau_k \in \mathcal{T'}} R_{\tau_k}\!\bigl(\hat{s}_{t}^{\text{env}}, \mathcal{A}_{t}\bigr) + R_{\text{env}}\!\bigl(\hat{s}_{t}^{\text{env}}\bigr),
\end{align}
where $\hat{s}_{t}^{\text{env}}$ denotes the rolled-out traffic state, and $R_{\tau_k}(\cdot)$ and $R_{\text{env}}(\cdot)$ evaluate task-specific and system-level performance. Evaluating generated actions in the unified environment reveals how local interventions benefit subsystems and affect system-level performance, which enables decision-making through an analysis--execution feedback loop.

\subsection{Spatiotemporal Memory Management}

\subsubsection{Episodic Spatiotemporal Context Cache}

Analytics from executable spatiotemporal reasoning are preserved within an episode using an Episodic Spatiotemporal Context Cache (ESCC), a structured working memory over spatial, temporal, and task dimensions that supports multi-step planning and reuse of analytical context.

During decision-making, the agent queries $\text{ESCC}_{t-1}$ to retrieve historical analytics that can be utilized in the current reasoning episode. The retrieved context is then combined with the current analytical result $Y_t^{\mathrm{ana}}$ for action generation:
\begin{equation}
\mathcal{A}_t = \Phi_{\text{act}}\!\left(Y_t^{\mathrm{ana}}, \text{Retrieve}(\text{ESCC}_{t-1}), \mathcal{T}'\right),
\end{equation}
where $\text{Retrieve}(\cdot)$ denotes the retrieval process of the agent to select relevant cached analytics. ESCC therefore grounds decisions in both the current diagnosis and prior context within the same episode.

\subsubsection{Procedural Spatiotemporal Memory}

While ESCC supports data-grounded reasoning within a single episode, long-horizon optimization requires accumulating experience across episodes. We introduce a Procedural Spatiotemporal Memory (PSM) that distills episode-level experience into persistent procedural knowledge. Instead of storing raw intermediate results, PSM abstracts recurring traffic patterns, coordination strategies, and failure modes, enabling continual self-improvement of reasoning and control during iterative interactions.

We perform an episode-level summarization step that extracts higher-level procedural insights, including: (1) \emph{recurring congestion patterns} (\eg peak-direction spillback chains or station overload), (2) \emph{effective system-level coordination strategies} (\eg cross-subsystem optimization), and (3) \emph{systematic failure modes} (\eg repeated spillback caused by locally myopic interventions). Formally, the episode-level summarization is represented as:
\begin{equation}
Y_e^{\mathrm{sum}} = \Phi_{\text{sum}}(\text{ESCC}_e, \mathcal{T}'),
\end{equation}
where $\Phi_{\text{sum}}(\cdot)$ denotes the summarization process of the LLM agent for extracting procedural insights from reasoning episodes, capturing recurring traffic patterns, coordination strategies, and failure modes within interactive episodes.

All episodic summaries are maintained in the persistent Procedural Spatiotemporal Memory (PSM), which aggregates long-horizon experience across episodes. To prevent uncontrolled memory growth while preserving representative and informative knowledge, the agent performs memory management through selective insertion, merging, and pruning. Given the newly abstracted insight $Y_e^{\mathrm{sum}}$ from episode $e$, the PSM is updated as:
\begin{equation}
\text{PSM}_e = \Phi_{\text{update}}(\text{PSM}_{e-1}, Y_e^{\mathrm{sum}}, \mathcal{T}'),
\end{equation}
where $\Phi_{\text{update}}(\cdot)$ denotes the reasoning process to update PSM. Similar items are merged, obsolete or low-value entries are pruned, and informative patterns are retained as long-term procedural knowledge for stable coordinated control.

\subsection{RL for System-Level Traffic Control}


\subsubsection{Supervised Multi-Task Cold-Start}

To bootstrap generalizable reasoning across heterogeneous subsystems and traffic dynamics, we first train the agent using supervised learning on diverse interaction trajectories. Training scenarios cover: (1) \textit{spatial regions}, spanning regions with different traffic networks; (2) \textit{subsystem task configurations}, including diverse suites of control tasks from heterogeneous traffic subsystems; and (3) \textit{temporal dynamics}, capturing daily traffic patterns including peak and off-peak fluctuations.

Reference trajectories are generated using a generalist LLM (\eg DeepSeek \cite{liu2025deepseek}), producing high-quality interaction traces under coupled cross-subsystem dynamics. To bias learning toward globally beneficial behaviors, we retain only trajectories that outperform optimization-based baselines in system-level traffic efficiency (\ie global travel time). The LLM backbone is then trained via supervised fine-tuning to imitate these high-quality interactions, providing a strong policy prior and stabilizing early-stage exploration for subsequent RL training.

\subsubsection{System-Aware Reward Construction}

To enable unified optimization under heterogeneous and tightly coupled objectives, we design a system-level reward that encourages the agent to optimize overall traffic efficiency rather than isolated subsystem-specific outcomes:
\vspace{-3pt}
\begin{align}
    &R_{\text{env}}(s_t, \mathcal{A}_{t:t+H})
    = \\
    & \sum_{\tau=t}^{t+H} f_{\text{TP}}(s_\tau, a_\tau) -f_{\text{TT}}(s_\tau, a_\tau) + f_{\text{RI}}(s_\tau, a_\tau) \notag
\end{align} 
where $f_{\text{TP}}(\cdot)$, $f_{\text{TT}}(\cdot)$, $f_{\text{RI}}(\cdot)$ denote normalized throughput, average travel time, and relative improvement over the optimization-based baseline, respectively. These signals quantify whether control actions across heterogeneous traffic subsystems contribute to system-level traffic efficiency.

Quantitative metrics alone may not capture whether decisions are behaviorally coherent across subsystems. We further introduce an extra coordination reward, instantiated through LLM-as-a-judge \cite{tan2025judgebench}, to evaluate cross-subsystem trade-offs, detect conflicting actions, and encourage alignment with system-level objectives $R_{\text{coord}}(s_t, \mathcal{A}_{t:t+H}, \mathcal{T}^\prime)$, which scores coordination quality, action consistency, and global traffic efficiency. The total reward is a weighted sum of system-level metrics and coordination evaluation:
\vspace{-2pt}
\begin{align}
    &R(s_t, \mathcal{A}_{t:t+H}, \mathcal{T}^\prime) = \\
    & R_{\text{env}}(s_t, \mathcal{A}_{t:t+H}) + R_{\text{coord}}(s_t, \mathcal{A}_{t:t+H}, \mathcal{T}^\prime). \notag
\end{align}

\subsubsection{System-Aligned Policy Optimization}

We optimize the agent with agentic RL using Group Relative Policy Optimization (GRPO)~\cite{guo2025deepseek}. The agent iteratively interacts with the traffic environment, observes how interventions propagate across subsystems, and updates its policy to maximize system-level traffic efficiency.

Formally, the agent policy is defined as $\pi_\theta(\mathcal{A} \mid \mathcal{O})$. The agent’s objective is to maximize the expected cumulative system-level reward $R$. GRPO updates by computing a advantage function $\text{Adv}_t$ and performing policy gradient:
\vspace{-3pt}
\begin{align}
    \theta \leftarrow \theta + \eta \, \mathbb{E}\big[ \nabla_\theta \log \pi_\theta(\mathcal{A} \mid \mathcal{O}) \cdot \text{Adv}_t\big],
\end{align}
Through multi-turn interaction, the agent internalizes how local interventions shape the global traffic state and learns robust system-level policies rather than isolated short-term gains.


\begin{table}[t]
  \centering
  \setlength{\tabcolsep}{2pt}
  \caption{Statistics of our studied regions, including road network scale and transportation demand dynamics.}
  \label{tab:env-statistics}
  \small
  \resizebox{\columnwidth}{!}{
  \begin{tabular}{l|ccc|cccc}
  \toprule
  \multirow{2}{*}{\textbf{Region}} & \multicolumn{3}{c}{\textbf{Road Network}} & \multicolumn{4}{c}{\textbf{Transportation Demand}} \\
  \cmidrule(lr){2-4}\cmidrule(lr){5-8}
  & Junctions & Lanes & Lane-km & Taxi & Public Transit & Walk & Total \\
  \midrule
  Manhattan & 10593 & 26329 & 1501.00 & 12111 & 51162 & 16816 & 98819 \\
  Queens & 17896 & 50461 & 2672.99 & 3748 & 45711 & 13105 & 86486 \\
  \bottomrule
  \end{tabular}}
  \vspace{-10pt}
\end{table}

\begin{table*}[t]
  \centering
  \small
  \setlength{\tabcolsep}{1pt}
  \caption{Experimental results across two NYC regions. Throughput (Thpt.), income, and completed trips (Trip) are higher-is-better ($\uparrow$), while travel time (Travel), fuel consumption (Fuel), queue length (Queue), waiting time (Wait), and electricity usage (Elec.) are lower-is-better ($\downarrow$).}
  \label{tab:main_results}
  \resizebox{\textwidth}{!}{
  \begin{tabular}{l|cc|cc|cc|cc|cc|cc|cc|cc|cc|cc}
  \toprule
  \multirow{6}{*}{\textbf{Method/Model}}
  & \multicolumn{10}{c|}{\textbf{Manhattan}}
  & \multicolumn{10}{c}{\textbf{Queens}} \\
  \cmidrule(lr){2-11}
  \cmidrule(lr){12-21}
  & \multicolumn{6}{c|}{in-domain}
  & \multicolumn{4}{c|}{out-of-domain}
  & \multicolumn{6}{c|}{in-domain}
  & \multicolumn{4}{c}{out-of-domain} \\
  \cmidrule(lr){2-7}\cmidrule(lr){8-11}\cmidrule(lr){12-17}\cmidrule(lr){18-21}
  & \multicolumn{2}{c|}{Signal Control}
  & \multicolumn{2}{c|}{Bus Sched.}
  & \multicolumn{2}{c|}{Taxi Dispatching}
  & \multicolumn{2}{c|}{Subway Sched.}
  & \multicolumn{2}{c|}{Ramp Metering}
  & \multicolumn{2}{c|}{Signal Control}
  & \multicolumn{2}{c|}{Bus Sched.}
  & \multicolumn{2}{c|}{Taxi Dispatching}
  & \multicolumn{2}{c|}{Subway Sched.}
  & \multicolumn{2}{c}{Ramp Metering} \\
  \cmidrule(lr){2-3}\cmidrule(lr){4-5}\cmidrule(lr){6-7}\cmidrule(lr){8-9}\cmidrule(lr){10-11}
  \cmidrule(lr){12-13}\cmidrule(lr){14-15}\cmidrule(lr){16-17}\cmidrule(lr){18-19}\cmidrule(lr){20-21}
  & Thpt. ($\uparrow$) & Travel ($\downarrow$)
  & Fuel ($\downarrow$) & Wait ($\downarrow$)
  & Income ($\uparrow$) & Trip ($\uparrow$)
  & Elec. ($\downarrow$) & Wait ($\downarrow$)
  & Travel ($\downarrow$) & Queue ($\downarrow$)
  & Thpt. ($\uparrow$) & Travel ($\downarrow$)
  & Fuel ($\downarrow$) & Wait ($\downarrow$)
  & Income ($\uparrow$) & Trip ($\uparrow$)
  & Elec. ($\downarrow$) & Wait ($\downarrow$) 
  & Travel ($\downarrow$) & Queue ($\downarrow$) \\
  \midrule
  Classic Methods
  & \textbf{122.07} & \textbf{481.80} & \textbf{339.69} & \textbf{512.11} & \textbf{6803.06} & \underline{590} & \underline{330.04} & \underline{603.01} & \textbf{138.16} & \textbf{4.08}
  & \textbf{152.52} & \textbf{576.38} & \textbf{669.81} & \textbf{462.60} & \underline{1925.61} & \textbf{300} & \underline{185.11} & \underline{761.33} & \textbf{148.21} & \textbf{11.46} \\
  RL-based Models
  & \underline{89.30} & \underline{527.98} & \underline{657.83} & \underline{539.69} & \underline{6485.59} & \textbf{610} & \textbf{231.32} & \textbf{340.26} & \underline{189.81} & \underline{4.86} 
  & \underline{100.12} & \underline{654.18} & \underline{1381.06} & \underline{509.44} & \textbf{2116.13} & \underline{282} & \textbf{170.71} & \textbf{242.62} & \underline{297.08} & \underline{12.40} \\
  \midrule
  \multicolumn{21}{c}{\textbf{LLM Agents}} \\
  \midrule
  TrafficGPT & 124.77 & \textbf{481.32} & 749.69 & \textbf{498.83} & 6901.45 & 725 & \underline{314.76} & 678.61 & \textbf{102.99} & \underline{4.29}
  & \underline{158.50} & \underline{570.57} & 1170.44 & 564.85 & \underline{3451.62} & \textbf{371} & \underline{218.45} & 391.12 & 296.66 & 12.73 \\
  SUMO-MCP & \underline{126.25} & 493.05 & \underline{702.45} & \underline{504.87} & \textbf{9463.35} & \underline{727} & 315.41 & \textbf{294.61} & \underline{106.20} & \textbf{3.53} 
  & \textbf{159.65} & \textbf{564.06} & \textbf{903.04} & \underline{471.71} & 2948.33 & 314 & 232.37 & \underline{216.26} & \underline{163.27} & \textbf{11.91} \\
  AIDE & \textbf{126.46} & \underline{491.50} & \textbf{257.58} & 511.95 & \underline{7945.27} & \textbf{777} & \textbf{276.31} & \underline{425.14} & 136.25 & 4.69 
  & 154.38 & 577.23 & \underline{1042.66} & \textbf{449.21} & \textbf{3571.23} & \underline{369} & \textbf{174.99} & \textbf{195.32} & \textbf{149.85} & \underline{12.58} \\
  \midrule
  \multicolumn{21}{c}{\textbf{TrafficClaw with Generalist LLMs}} \\
  \midrule
  DeepSeek-V3.2 & 121.98 & \underline{493.48} & \underline{927.15} & 419.42 & 7375.28 & 756 & \underline{260.82} & 358.27 & 130.65 & \textbf{3.21} 
  & 152.35 & 576.06 & \underline{1668.97} & \underline{458.92} & \textbf{3571.62} & 315 & 292.05 & 367.36 & 206.91 & 12.97 \\
  MiniMax-M2.5 & \textbf{127.04} & 493.84 & 1476.15 & 479.56 & \underline{8849.25} & \underline{767} & 291.00 & \underline{324.92} & 124.93 & 4.07 
  & \underline{160.02} & \textbf{566.78} & 2282.35 & 468.99 & 1652.83 & 260 & \textbf{155.75} & 722.67 & 153.07 & \textbf{11.75} \\
  Kimi-K2.5 & \underline{126.25} & 497.24 & 1195.44 & \textbf{391.98} & 7752.26 & 536 & \textbf{250.58} & 357.27 & 140.50 & \underline{3.34} 
  & 152.33 & 572.12 & 2518.40 & 488.69 & 1921.63 & \textbf{329} & 249.85 & \underline{312.08} & \textbf{95.09} & 13.23 \\
  Gemini-3.1-Pro & 123.31 & 494.22 & 1006.75 & \underline{398.29} & 7183.84 & 608 & 287.90 & 387.61 & \textbf{81.38} & 3.83 
  & 148.60 & 580.72 & 1880.37 & \textbf{441.01} & 1524.42 & 225 & 215.54 & \textbf{213.34} & 165.81 & \underline{12.08} \\
  o4-mini & 125.00 & \textbf{489.58} & \textbf{800.09} & 434.84 & \textbf{9787.97} & \textbf{837} & 352.94 & \textbf{259.82} & \underline{103.29} & 3.80 
  & \textbf{166.13} & \underline{567.16} & \textbf{996.86} & 499.39 & \underline{3411.14} & \underline{318} & \underline{208.90} & 392.94 & \underline{126.31} & 12.62 \\
  \midrule
  \multicolumn{21}{c}{\textbf{TrafficClaw with Base Model Family}} \\
  \midrule
  Qwen3-8B & \underline{124.17} & 487.60 & 307.30 & 502.74 & 9201.60 & 781 & 317.08 & \underline{350.92} & \underline{99.77} & 3.76 
  & 138.67 & 584.25 & \textbf{686.50} & 507.52 & \underline{3139.57} & \underline{315} & 209.41 & 286.84 & 108.29 & \textbf{12.22} \\
  Qwen3-32B & 121.13 & 488.68 & \underline{295.91} & 532.29 & 6055.99 & 524 & \textbf{258.85} & 412.06 & 127.86 & \textbf{3.38} 
  & 149.19 & 583.74 & \underline{793.40} & 528.89 & 1645.00 & 248 & 194.89 & 313.98 & 141.16 & \underline{12.38} \\
  Qwen3-235B & 123.13 & \underline{486.25} & 429.80 & \underline{490.24} & \underline{9327.90} & \underline{873} & \underline{270.22} & 427.45 & 108.75 & 3.68 & 
  \underline{150.73} & \textbf{575.61} & 980.78 & \underline{494.84} & 1848.09 & 244 & \textbf{163.48} & \underline{249.90} & \underline{103.13} & 12.54 \\
  TrafficClaw (8B) & \textbf{126.04} & \textbf{482.97} & \textbf{256.16} & \textbf{437.17} & \textbf{11017.49} & \textbf{1149} & 294.53 & \textbf{220.99} & \textbf{96.09} & \underline{3.64} 
  & \textbf{159.69} & \underline{578.91} & 823.68 & \textbf{450.12} & \textbf{3781.19} & \textbf{486} & \underline{194.23} & \textbf{177.03} & \textbf{100.51} & 14.16 \\
  \bottomrule
  \end{tabular}}
  \vspace{-10pt}
\end{table*}

\begin{table}[t]
  \centering
  \caption{Transferability comparison in Brooklyn.}
  \label{tab:transferability_comparison}
  \small
  \setlength{\tabcolsep}{2pt}
  \resizebox{\columnwidth}{!}{%
  \begin{tabular}{l*{6}{c}}
      \toprule
      \multirow{2}{*}{\textbf{Method/Model}} & \multicolumn{2}{c}{\textbf{Signal Control}} & \multicolumn{2}{c}{\textbf{Bus Scheduling}} & \multicolumn{2}{c}{\textbf{Taxi Dispatching}} \\
      \cmidrule(lr){2-3}\cmidrule(lr){4-5}\cmidrule(lr){6-7}
      & Thpt. ($\uparrow$) & Travel ($\downarrow$) & Fuel ($\downarrow$) & Travel ($\downarrow$) & Income ($\uparrow$) & Trip ($\uparrow$) \\
      \midrule
      Classic Method & 160.71 & 509.55 & 226.34 & \underline{231.18} & 748.30 & 99 \\
      RL-based Model & 125.83 & 564.19 & 362.09 & \textbf{205.32} & \underline{845.01} & 106 \\
      \midrule
      DeepSeek-V3.2 & 158.71 & \textbf{498.26} & 173.19 & 268.85 & 840.19 & \underline{109} \\
      MiniMax-M2.5 & \textbf{161.21} & 503.46 & 171.65 & 311.37 & 726.02 & 99 \\
      Kimi-K2.5 & 160.38 & 505.76 & \textbf{148.92} & 273.30 & 775.26 & 104 \\
      o4-mini & \underline{161.04} & 518.26 & 172.89 & 315.83 & 675.75 & 88 \\
      \midrule
      TrafficClaw (8B) & 160.25 & \underline{500.26} & \underline{150.83} & 238.07 & \textbf{976.76} & \textbf{123} \\
      \bottomrule
  \end{tabular}}
  \vspace{-10pt}
\end{table}

\section{Experiments}

We conduct extensive experiments to evaluate TrafficClaw by answering these research questions:
\begin{itemize}[leftmargin=0.3cm,itemsep=0pt,topsep=0.2em,parsep=0pt]
    \item \textbf{RQ1}: How does TrafficClaw compare to the state-of-the-art traffic control agents in effectiveness and generalization ability?
    \item \textbf{RQ2}: How is TrafficClaw's system-level optimization performance across subsystems?
    \item \textbf{RQ3}: How does each component of TrafficClaw improve its reasoning and decision-making?
\end{itemize}

\subsection{Experimental Settings}

\noindent \textbf{Environment Settings}: Our experimental regions include Manhattan, Queens, and Brooklyn in New York City, USA, three metropolitan regions with different spatial scales and demand dynamics (Table~\ref{tab:env-statistics}). We use SUMO~\cite{alvarez_lopez_2026_18406080} for simulation, covering subsystems of traffic signals, freeways, public transit, and taxi services. The simulation spans 24 hours, including rush hours \cite{adeyemi2021exploring} from 6:00 am to 10:00 am and 3:00 pm to 8:00 pm. The details of selected tasks are summarized in Appendix~\ref{app:task-definitions}. The environment exposes 98 dynamic traffic-state features, 20 static data types, and 15 interaction APIs spanning subsystem and global operations. The details of the data interface are in Appendix~\ref{app:env-api-details}.

\vspace{3pt}
\noindent \textbf{Baseline Methods}: Baselines include classic methods (\eg Webster \cite{koonce2008traffic}), RL-based models (\eg CoLight~\cite{wei2019colight}), traffic LLM agents (TrafficGPT~\cite{zhang2024trafficgpt}, SUMO-MCP~\cite{ye2025sumomcp}, and general data modeling agent AIDE~\cite{jiang2025aide}), and leading LLMs (Qwen3~\cite{yang2025qwen3}, DeepSeek-V3.2~\cite{liu2025deepseek}, MiniMax-M2.5~\cite{minimax_m2_2025}, Kimi-K2.5~\cite{team2026kimi}, o4-mini~\cite{openai_o3_o4_mini_2025}, and Gemini-3.1-Pro~\cite{deepmind_gemini}). Details are in Appendix~\ref{app:baseline-details}.

\vspace{3pt}
\noindent \textbf{Training and Evaluation Settings}: Qwen3-8B is used as the base model. In training, the agent is exposed to signal control, highway speed limit control, bus scheduling, and taxi dispatching, along with their corresponding cooperation settings. To assess generalization, ramp metering, subway scheduling, and their associated cooperative scenarios are held out for evaluation. Manhattan and Queens are used for training, while Brooklyn is reserved for evaluating cross-region transferability.

\subsection{Overall Performance Comparison (RQ1)}

Table~\ref{tab:main_results} reports results on Manhattan and Queens under both in-domain and out-of-domain settings. Traditional optimization methods remain competitive on rule-driven objectives but struggle in heterogeneous subsystems due to limited adaptability to complex urban dynamics. RL-based approaches require substantial task-specific engineering and scale poorly in city-wide environments with large state and action spaces. They only achieve competitive performance mainly on low-dimensional tasks, such as subway scheduling with only a few controllable lines. Traffic LLM agents (\ie TrafficGPT and SUMO-MCP) perform well on traffic signal control but degrade substantially on other subsystems, revealing limited cross-task generalization. In contrast, TrafficClaw consistently matches or surpasses leading LLM baselines despite using only an 8B backbone, while achieving more balanced trade-offs across conflicting objectives, such as reducing bus waiting time without excessive energy consumption. Table~\ref{tab:transferability_comparison} further shows that TrafficClaw transfers effectively to Brooklyn while maintaining stable system-level performance.

\begin{figure}[t]
  \centering
  \includegraphics[width=\columnwidth]{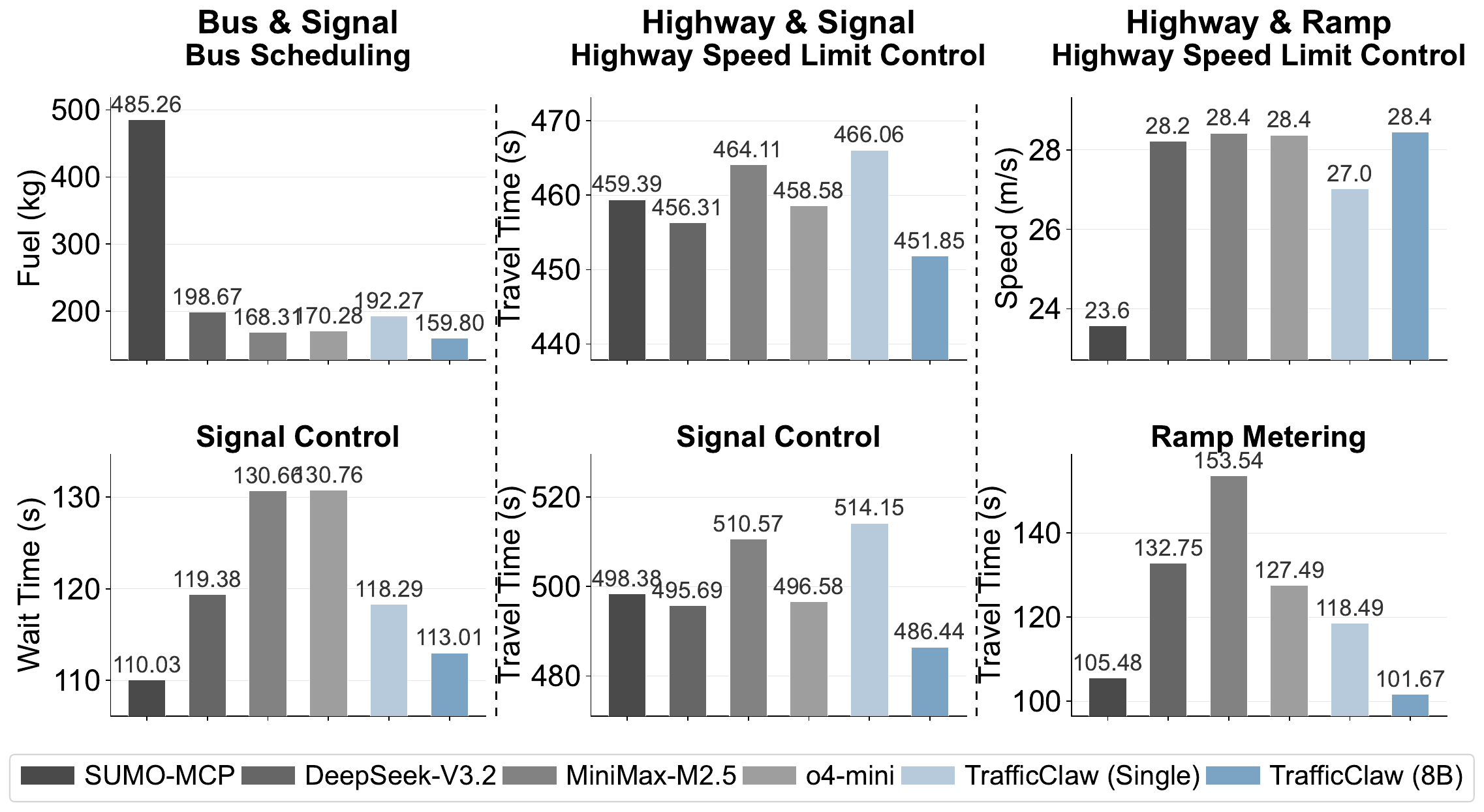}
  \caption{Multi-task cooperation comparison.}
  \Description{Multi-panel chart comparing cooperative multi-task traffic control performance across methods for bus--signal, highway--signal, and highway--ramp task suites.}
  \label{fig:cooperative_comparison}
  \vspace{-15pt}
\end{figure}

\subsection{Cross-Subsystem Coordination (RQ2)}\label{subsec:coordination}

We evaluate cooperative control on Manhattan across three cross-subsystem task suites: bus scheduling with signal control, highway speed limit control with signal control, and highway speed limit control with ramp metering over a 12-hour period. As shown in Figure~\ref{fig:cooperative_comparison}, TrafficClaw consistently achieves the strongest joint performance by optimizing coupled subsystems rather than isolated tasks. In the bus--signal setting, it minimizes bus fuel consumption while maintaining near-optimal signal delay, indicating that coordinated signals reduce bus stopping and queuing. In the highway-related settings, TrafficClaw simultaneously improves highway efficiency, urban travel time, and ramp accessibility, whereas baselines typically optimize one objective at the expense of others.

Compared with TrafficClaw (Single), which optimizes each subsystem independently, cooperative TrafficClaw achieves more balanced gains across coupled tasks. Table~\ref{tab:cooperative_global_travel_time} further demonstrates the importance of system-level coordination, with TrafficClaw consistently achieving the lowest global travel time. Furthermore, removing the coordination reward $R_{\text{coord}}$ in TrafficClaw (w/o Co-Reward) leads to consistently higher travel time. This suggests that our proposed coordination rewards help resolve cross-subsystem trade-offs and mitigate conflicting actions. Overall, jointly optimizing coupled subsystems yields substantially stronger network-wide traffic efficiency.

\begin{table}[t]
  \centering
  \caption{Average travel time (seconds) of all vehicles within the road network in Manhattan.}
  \label{tab:cooperative_global_travel_time}
  \small
  \setlength{\tabcolsep}{4pt}
  \resizebox{\columnwidth}{!}{%
  \begin{tabular}{lccc}
      \toprule
      \textbf{Method} & \textbf{Bus--Signal} & \textbf{Highway--Signal} & \textbf{Highway--Ramp} \\
      \midrule
      Classic Method & 486.65 & 462.73 & 512.97 \\
      RL-based Model & 499.10 & \underline{456.17} & 528.23 \\
      TrafficClaw (Single) & \underline{465.95} & 459.71 & \underline{505.04} \\
      TrafficClaw (w/o Co-Reward) & 461.06 & 463.40 & 506.06 \\
      TrafficClaw & \textbf{459.87} & \textbf{451.85} & \textbf{504.62} \\
      \bottomrule
  \end{tabular}}
  \vspace{-15pt}
\end{table}

\subsection{Ablation Studies (RQ3)}\label{subsec:ablation}

\begin{figure}[t]
  \centering
  \includegraphics[width=\columnwidth]{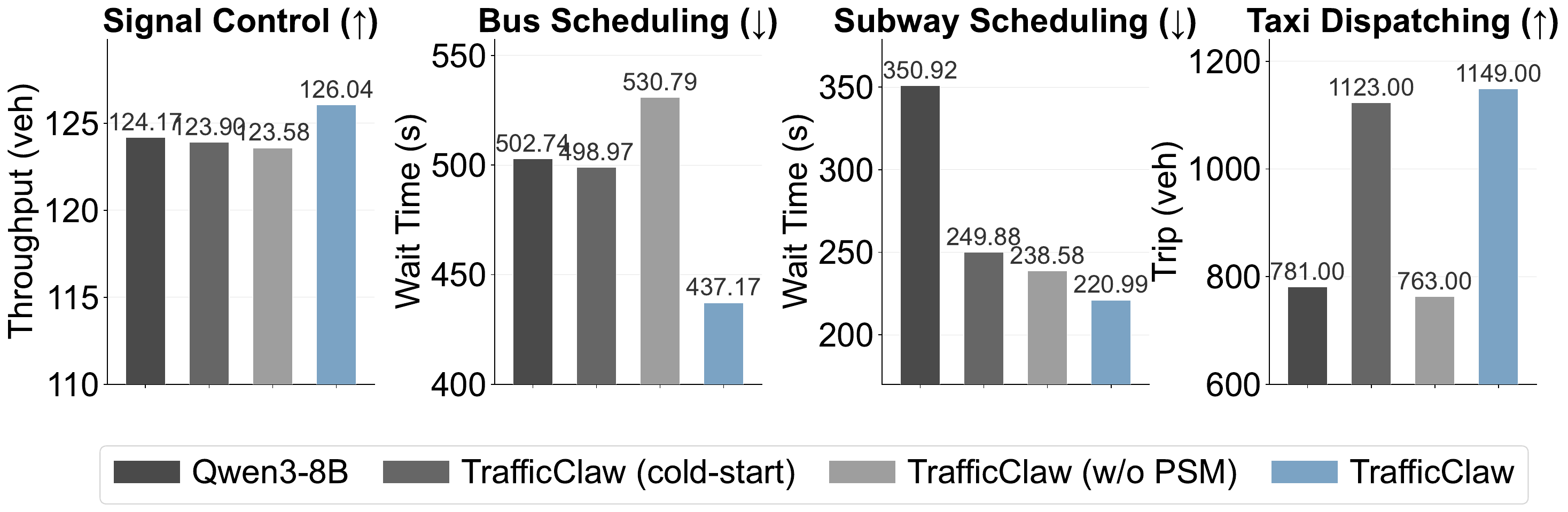}
  \caption{Ablation study on TrafficClaw.}
  \Description{Chart summarizing ablation results when removing procedural spatiotemporal memory or RL tuning across multiple traffic control metrics.}
  \label{fig:ablation}
  \vspace{-10pt}
\end{figure}

As shown in Figure~\ref{fig:ablation}, supervised cold-start already improves substantially over the base model but remains unbalanced performance, indicating that imitation alone cannot capture cross-subsystem dynamics. RL-based tuning further refines policies through trial-and-error learning, yielding better system-level efficiency. Removing PSM causes systematic degradation across tasks, indicating that PSM is crucial for long-horizon adaptation and self-improvement. Appendix~\ref {app:memory-case} further shows that PSM increasingly improves performance over time by accumulating reusable procedural knowledge through iterative interactions.

\section{Related Work}

\noindent \textbf{LLM Agents}:
LLM agents extend language models with planning, tool use, environmental interaction, and episodic memory~\cite{li2025webthinker,pang2025browsemaster,dong2024villageragent,openclaw2025}. Recent urban agents further introduce spatiotemporal reasoning~\cite{feng2025agentmove,li_stbench_2025,lai_ustbench_2026}, but they are mainly evaluated in loosely coupled settings rather than multi-subsystem traffic environments with shared physical constraints.

\vspace{5pt}
\noindent \textbf{Traffic Control}:
Traffic control optimizes efficiency and sustainability \cite{tian2025arrow,tian2024air-dualode} of urban traffic systems. It has evolved from classical optimization and RL toward LLM-based methods~\cite{wei2019survey,siri2021freeway,farazi2021deep}. LLM agents incorporate reasoning and flexible tool use for decision-making~\cite{han2025large}. However, existing approaches remain largely task-specific and cannot effectively model coupled dynamics across heterogeneous traffic subsystems.

\vspace{5pt}
\noindent \textbf{Agentic Reinforcement Learning}:
Agentic RL improves LLM decision-making through interaction, feedback, tool use, and long-horizon trajectories~\cite{guo2025deepseek,fang2025towards,hu2025agentgen,liang2024environment,froger2025scaling,xi2025agentgym}. However, most studies target digital environments, whereas urban traffic requires system-level optimization across heterogeneous and sometimes conflicting objectives.
\section{Conclusion}

We introduce TrafficClaw as a step toward physically grounded LLM agents for urban systems. Instead of treating traffic control as isolated tasks, TrafficClaw embeds agents within a unified physical traffic environment where shared infrastructure, mobility demand, and local interventions jointly shape network-wide dynamics. By integrating executable spatiotemporal reasoning, persistent memory, and system-level agentic RL, TrafficClaw enables long-horizon coordinated control across heterogeneous subsystems. Our results suggest that physical-world LLM agents require not only reasoning abilities, but also a unified environment that exposes coupled dynamics and feedback-driven optimization for system-level coordination.

\clearpage
\section*{Limitations}

TrafficClaw focuses on unified agentic traffic control within large-scale urban environments. While the framework demonstrates strong generalization across regions, subsystems, and unseen task settings, several promising directions remain for future work. First, extending TrafficClaw to real-world deployment with online traffic streams, connected infrastructure, and human-in-the-loop operations would further validate its practicality under dynamic urban conditions. Second, future research may incorporate richer mobility behaviors, long-term traveler adaptation, and additional urban objectives such as emergency response and accessibility. Finally, integrating stronger safety constraints, uncertainty-aware reasoning, and more robust memory and reward mechanisms could further improve the reliability and scalability of long-horizon autonomous traffic control.

\section*{Ethical Considerations}

TrafficClaw is designed to support more efficient and coordinated urban mobility management. Our experiments rely on aggregate, publicly available traffic and mobility data and do not require personally identifiable information. Nevertheless, traffic-control systems can influence congestion, accessibility, and service quality across different regions and populations. Real-world deployment should therefore include evaluation of fairness, environmental impact, and robustness under diverse operating conditions. We view TrafficClaw as a research and decision-support framework rather than a replacement for transportation authorities. Practical deployment should incorporate human oversight, transparent logging, safety constraints, and conservative rollout procedures to ensure accountable and reliable operation in real-world urban systems.

\bibliography{references}

\appendix
\clearpage

\begin{table*}[t]
    \centering
    \caption{Detailed dynamic traffic-state features.}
    \label{tab:appendix-observation-features}
    \scriptsize
    \setlength{\tabcolsep}{3pt}
    \renewcommand{\arraystretch}{0.96}
    \begin{tabularx}{\textwidth}{p{0.12\textwidth}|X|p{0.04\textwidth}|p{0.21\textwidth}}
    \toprule
    \textbf{Category} & \textbf{Feature Names} & \textbf{Count} & \textbf{Description} \\
    \midrule
    \textbf{Lane-level Features} &
    \texttt{queue\_length}, \texttt{queue\_density}, \texttt{moving\_vehicles}, \texttt{average\_speed}, \texttt{average\_waiting\_time}, \texttt{cell\_occupancy}, \texttt{lane\_density}, \texttt{throughput\_potential}, \texttt{occupancy}, \texttt{halting\_number}, \texttt{max\_speed}, \texttt{arrival\_rate}, \texttt{entering\_vehicles}, \texttt{vehicle\_count}, \texttt{vehicle\_details} (\texttt{speed}, \texttt{position}, \texttt{waiting\_time}) & 
    15 &
    Lane-level traffic-state features returned by \texttt{read\_lane\_traffic\_states}. \\
    \midrule
    \textbf{Highway Features} &
    \texttt{segment\_speed}, \texttt{segment\_density}, \texttt{segment\_occupancy}, \texttt{segment\_speed\_limit}, \texttt{segment\_default\_speed\_limit}, \texttt{segment\_congestion\_ratio}, \texttt{segment\_speed\_ratio}, \texttt{segment\_speed\_pressure}, \texttt{road\_speed}, \texttt{road\_density}, \texttt{road\_occupancy}, \texttt{current\_speed\_limits}, \texttt{default\_speed\_limits} &
    13 &
    Freeway segment-level traffic-state features returned by \texttt{read\_highway\_traffic\_ states}. \\
    \midrule
    \textbf{Ramp Features} &
    \texttt{vehicle\_count}, \texttt{queue\_length}, \texttt{queue\_density}, \texttt{moving\_vehicles}, \texttt{average\_speed}, \texttt{average\_waiting\_time}, \texttt{cell\_occupancy}, \texttt{lane\_density}, \texttt{occupancy}, \texttt{halting\_number}, \texttt{max\_speed}, \texttt{arrival\_rate}, \texttt{lane\_length}, \texttt{road\_id}, \texttt{direction}, \texttt{start\_intersection}, \texttt{end\_intersection} &
    17 &
    Ramp-lane traffic-state features returned by \texttt{read\_ramp\_lane\_traffic\_ states}. \\
    \midrule
    \textbf{Bus Features} &
    \texttt{active\_buses}, \texttt{headway}, \texttt{station\_count}, \texttt{departure\_time}, \texttt{travel\_time}, \texttt{current\_edge}, \texttt{speed}, \texttt{passenger\_count}, \texttt{capacity}, \texttt{load\_ratio}, \texttt{next\_station}, \texttt{next\_station\_dwell\_time}, \texttt{waiting\_count}, \texttt{avg\_waiting\_time}, \texttt{max\_waiting\_time}, \texttt{waiting\_time\_distribution} (0--60s, 60--180s, 180--300s, $>$300s) &
    16 &
    Bus operation features returned by \texttt{read\_bus\_states}. \\
    \midrule
    \textbf{Subway Features} &
    \texttt{active\_trains}, \texttt{headway}, \texttt{station\_count}, \texttt{departure\_time}, \texttt{travel\_time}, \texttt{current\_edge}, \texttt{speed}, \texttt{passenger\_count}, \texttt{capacity}, \texttt{load\_ratio}, \texttt{next\_station}, \texttt{next\_station\_dwell\_time}, \texttt{waiting\_count}, \texttt{avg\_waiting\_time}, \texttt{max\_waiting\_time}, \texttt{waiting\_time\_distribution} (0--60s, 60--180s, 180--300s, $>$300s) &
    16 &
    Subway operation features returned by \texttt{read\_subway\_states}. \\
    \midrule
    \textbf{Taxi Features} &
    \texttt{fleet\_size}, \texttt{idle\_count}, \texttt{pickup\_count}, \texttt{occupied\_count}, \texttt{utilization\_rate}, \texttt{pending\_reservations}, \texttt{taxi\_state}, \texttt{customers}, \texttt{current\_edge}, \texttt{current\_taz}, \texttt{position}, \texttt{speed}, \texttt{cumulative\_income}, \texttt{recent\_order\_count} &
    14 &
    Taxi fleet features returned by \texttt{read\_taxi\_traffic\_states} and \texttt{taxi\_fleet\_state}. \\
    \midrule
    \textbf{Global Features} &
    \texttt{total\_vehicles}, \texttt{avg\_queue\_length}, \texttt{avg\_speed}, \texttt{avg\_waiting\_time}, \texttt{congestion\_level}, \texttt{intersection\_count}, \texttt{lane\_count} &
    7 &
    Global aggregated features returned by \texttt{analyze\_zone\_traffic}. \\
    \bottomrule
    \end{tabularx}
\end{table*}

\begin{table*}[t]
    \centering
    \caption{Detailed static traffic-state data.}
    \label{tab:appendix-static-resources}
    \scriptsize
    \setlength{\tabcolsep}{4pt}
    \renewcommand{\arraystretch}{1.02}
    \begin{tabularx}{\textwidth}{p{0.33\textwidth}|p{0.16\textwidth}|X}
    \toprule
    \textbf{Data Resource Name} & \textbf{Type} & \textbf{Description} \\
    \midrule
    \texttt{zone\_dict} & \texttt{dict} & Zone infrastructure dictionary containing lanes, intersections, highways, ramps, and transit stations. \\
    \texttt{lane\_dict} & \texttt{dict} & Lane metadata dictionary containing location, direction, lane group, and related attributes. \\
    \texttt{lane\_inter\_graph} & NetworkX \texttt{DiGraph} & Directed graph connecting lane groups and intersections. \\
    \texttt{highway\_segment\_graph} & NetworkX \texttt{DiGraph} & Directed graph describing connectivity among freeway segments. \\
    \texttt{highway\_segment\_dict} & \texttt{dict} & Freeway segment metadata dictionary. \\
    \texttt{ramp\_lane\_graph} & NetworkX \texttt{DiGraph} & Directed graph describing ramp-lane topology and upstream/downstream relations. \\
    \texttt{transit\_graph} & NetworkX \texttt{DiGraph} & Public transit network graph covering bus and subway routes and stations. \\
    \texttt{zone\_graph} & NetworkX \texttt{DiGraph} & Directed graph describing zone adjacency relations. \\
    \texttt{network\_graphs} & \texttt{dict} & Base-layer network graphs, including \texttt{lane\_graph}, \texttt{road\_graph}, and \texttt{transit\_graph}. \\
    \texttt{network\_dicts} & \texttt{dict} & Base-layer metadata dictionaries, including \texttt{lane\_dict}, \texttt{road\_dict}, and \texttt{station\_dict}. \\
    \texttt{bus\_route\_info} & \texttt{dict} & Detailed bus route information, including static structure and runtime operation data. \\
    \texttt{current\_signal\_config} & \texttt{dict} & Current signal control configuration. \\
    \texttt{current\_highway\_speed\_limit\_config} & \texttt{dict} & Current freeway speed-limit control configuration. \\
    \texttt{current\_ramp\_metering\_config} & \texttt{dict} & Current ramp-metering control configuration. \\
    \texttt{current\_bus\_schedule} & \texttt{dict} & Current bus scheduling configuration. \\
    \texttt{current\_subway\_schedule} & \texttt{dict} & Current subway scheduling configuration. \\
    \texttt{current\_taxi\_config} & \texttt{dict} & Current taxi dispatch and fleet-management configuration. \\
    \texttt{taxi\_fleet\_state} & \texttt{dict} & Current taxi fleet state. \\
    \texttt{pending\_reservations} & \texttt{dict} & Pending taxi reservation requests awaiting assignment. \\
    \texttt{taz\_stats} & \texttt{dict} & TAZ-level statistics including demand, supply, and matching rate. \\
    \bottomrule
    \end{tabularx}
\end{table*}

\begin{table*}[t]
    \centering
    \caption{Detailed callable interaction APIs.}
    \label{tab:appendix-callable-functions}
    \footnotesize
    \setlength{\tabcolsep}{4pt}
    \renewcommand{\arraystretch}{1.1}
    \begin{tabularx}{\textwidth}{p{0.28\textwidth}|X}
    \toprule
    \textbf{Function Name} & \textbf{Function Description} \\
    \midrule
    \texttt{read\_bus\_states} & Query historical bus operation states, including vehicle conditions and station waiting information. \\
    \texttt{read\_highway\_traffic\_states} & Query historical freeway traffic states, including speed, density, and occupancy. \\
    \texttt{predict\_arima} & Perform ARIMA-based time-series forecasting for future traffic flow patterns (\eg future traffic patterns or downstream lane occupancy). \\
    \texttt{analyze\_zone\_traffic} & Analyze zone- or TAZ-level traffic conditions by aggregating lane-level indicators. \\
    \texttt{identify\_congestion\_hotspots} & Identify congestion hotspots in the network based on queue-length and speed thresholds. \\
    \texttt{calculate\_network\_metrics} & Compute network-level traffic indicators, such as congestion index, average speed, and throughput potential. \\
    \texttt{read\_ramp\_lane\_traffic\_states} & Query historical ramp-lane traffic states, including occupancy and queue length. \\
    \texttt{read\_lane\_traffic\_states} & Query historical lane traffic states, including queue length, speed, and waiting time. \\
    \texttt{read\_subway\_states} & Query historical subway operation states, including train conditions and station waiting information. \\
    \texttt{read\_taxi\_traffic\_states} & Query historical taxi traffic states, including fleet utilization and pending orders. \\
    \texttt{dispatch\_taxi} & Construct taxi dispatch decisions by assigning idle taxis to pending reservations. \\
    \texttt{reposition\_taxi} & Construct taxi repositioning decisions by moving idle taxis to target zones. \\
    \texttt{rank\_idle\_taxis\_by\_distance} & Rank idle taxis by Euclidean distance for distance-aware dispatching. \\
    \texttt{get\_zone\_infrastructure} & Query infrastructure information for a target zone. \\
    \texttt{get\_zones\_by\_infrastructure} & Retrieve zones containing specified infrastructure elements. \\
    \bottomrule
    \end{tabularx}
\end{table*}

\section{Observation and Interaction Interfaces}
\label{app:env-api-details}

Table~\ref{tab:appendix-observation-features}-\ref{tab:appendix-callable-functions} shows the dynamic traffic-state features, static data, and interaction APIs given to the agent.

\section{Evaluated Baselines}\label{app:baseline-details}

We summarize the baselines used in our comparisons below, moving the detailed descriptions out of the main text while keeping enough context to interpret their roles in the experiments. All LLM-based agents and baselines are provided with the same environment interaction interfaces as TrafficClaw for fair comparison. The training settings of RL-based models, including reward functions, hyperparameters, and input features, follow the configurations reported in the original papers.

\begin{itemize}[leftmargin=0.3cm, itemsep=0.35em plus 0.1em minus 0.05em, topsep=0.25em, parsep=0pt, partopsep=0pt]
    \item \textbf{Classic Method}: For classic methods, we use Webster \cite{koonce2008traffic} for signal control; a greedy heuristic \cite{sumo_taxi_doc} that drives taxis to the nearest pickup location for taxi dispatching; ALINEA~\cite{papageorgiou1991alinea} for ramp metering; the default policies published by NYC \cite{mta_open_data,osmplanet,nyc_mobility_survey} for bus and subway scheduling, highway speed limit control.
    \item \textbf{RL-based Model}: For RL-based models, we use CoLight~\cite{wei2019colight} for signal control; DQN \cite{mnih2013playing} for taxi dispatching; MARL \cite{wang2022robust} for bus scheduling; AutoDwell \cite{wang2022shortening} for subway scheduling; DDQN \cite{xu2022novel} for ramp metering; actor-critic RL \cite{wu2020differential} for highway speed limit control. Same as TrafficClaw, we use Manhattan and Queens for training, while Brooklyn is reserved for evaluating cross-region transferability. Epoch is set to 100, batch size is set to 64, learning rate is set to 0.0003, and reward discount factor is set to 0.95 for all tasks.
    \item \textbf{TrafficGPT}~\cite{zhang2024trafficgpt} is a traffic LLM agent that combines natural-language reasoning with traffic analytics tools for decision support.
    \item \textbf{SUMO-MCP}~\cite{ye2025sumomcp} is an autonomous traffic simulation agent built on the Model Context Protocol, enabling tool calling for scenario generation, simulation execution, and strategy comparison in SUMO.
    \item \textbf{AIDE}~\cite{jiang2025aide} is a general coding agent that searches over executable programs, making it a strong baseline for data-driven modeling and optimization.
    \item \textbf{Qwen3-Series}~\cite{yang2025qwen3} (8B, 32B, 235B) is developed by Alibaba Cloud, serves as the foundational open-source backbone in our training pipeline. We include it to assess the improvements achieved through our environment scaling and agentic reinforcement learning, rather than the backbone alone.
    \item \textbf{DeepSeek-V3.2}~\cite{liu2025deepseek} is an open-source generalist model that supports adaptive reasoning, which achieves competitive performance in reasoning and coding tasks.
    \item \textbf{MiniMax-M2.5}~\cite{minimax_m2_2025} is an open-source LLM for long-horizon agentic tasks.
    \item \textbf{Kimi-K2.5}~\cite{team2026kimi} is from Moonshot AI with strong reasoning and autonomous problem-solving capabilities.
    \item \textbf{o4-mini}~\cite{openai_o3_o4_mini_2025} is from OpenAI with strong tool-use and deliberate reasoning capabilities, offering an efficiency-oriented proprietary baseline.
    \item \textbf{Gemini-3.1-Pro}~\cite{deepmind_gemini} is a proprietary multimodal reasoning model with strong long-context and coding capabilities, included as another SOTA generalist baseline.
\end{itemize}

\section{Tasks and Evaluation Metrics}\label{app:task-definitions}

We summarize each evaluated task, including its goal in traffic control and the evaluation metrics, in the following list.

\begin{itemize}[leftmargin=0.3cm]
    \item \textbf{Signal control (traffic signals).} The agent optimizes signal phases by adjusting signal cycle time and phase durations to improve efficiency, evaluated by throughput (veh/h), average waiting time (s), and average travel time (s).
    \item \textbf{Highway speed limit control (freeways).} The agent adjusts variable speed limits to regulate mainline flow and avoid congestion, evaluated by average travel time and average speed (m/s).
    \item \textbf{Ramp metering (freeways).} The agent meters ramp by adjusting ramp open duration to protect mainline conditions while limiting ramp-side congestion, evaluated by average travel time and average queue length.
    \item \textbf{Bus scheduling (public transit).} The agent adjusts bus departures by adjusting bus stop dwell time and frequency to balance service reliability and operational cost, evaluated by fuel consumption (kg) and passenger waiting time (s).
    \item \textbf{Subway scheduling (public transit).} The agent adjusts subway departures by adjusting subway stop dwell time and frequency to balance passenger service and energy use, evaluated by electricity consumption (kWh) and waiting time.
\end{itemize}

\begin{figure}[t]
    \centering
    \includegraphics[width=0.9\columnwidth]{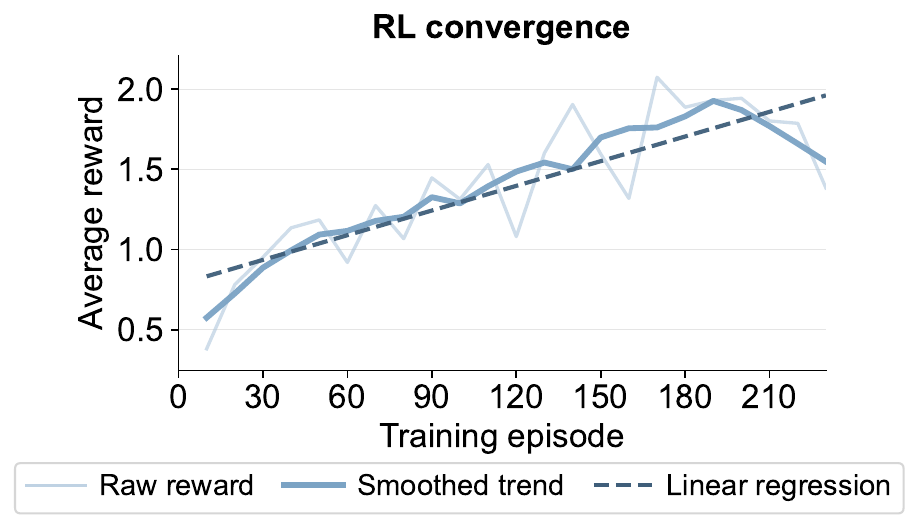}
    \caption{RL convergence of the agent during training.}
    \Description{Learning curve plotting average reward per episode versus training episodes, showing reward leveling off after roughly sixty episodes.}
    \label{fig:rl_convergence}
    \vspace{-10pt}
\end{figure}

\section{Agentic RL Convergence Analysis}\label{app:rl-convergence}

We analyze the agentic RL convergence during training. Figure~\ref{fig:rl_convergence} plots the average reward per episode. The reward stabilizes during training, indicating the effectiveness of our training.

\section{Memory Accumulation}\label{app:memory-case}

\begin{figure}[t]
  \centering
  \includegraphics[width=0.85\columnwidth]{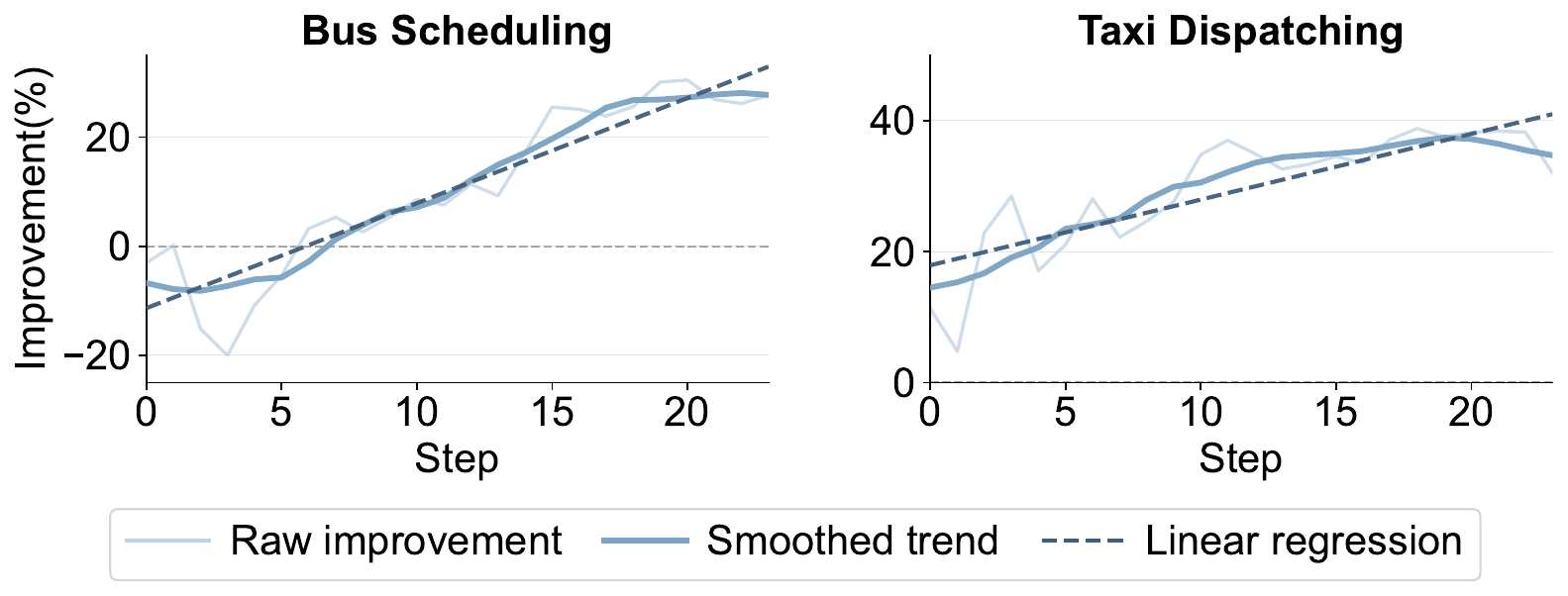}
  \caption{Improvement in agent control performance over time compared to TrafficClaw (w/o PSM).}
  \Description{Line or bar plot showing performance improving over interaction episodes with memory enabled versus without procedural spatiotemporal memory.}
  \label{fig:memory_accumulation_effect}
  \vspace{-10pt}
\end{figure}

Figure~\ref{fig:memory_accumulation_effect} shows the benefits of memory become increasingly pronounced over time as the agent accumulates experience through iterative interactions. This indicates that PSM serves as a mechanism for consolidating observed traffic dynamics and effective control strategies into reusable procedural knowledge. The key insight is that memory helps the agent accumulate, reuse, and refine effective coordination routines, thereby supporting self-improvement across traffic regimes and tasks.

\section{Case Study}\label{app:case-study}

Figure~\ref{fig:case_study_psm} presents two representative Manhattan control episodes that illustrate how PSM converts prior interaction experience into reusable procedural knowledge. PSM stores structured coordination patterns that capture cross-subsystem dependencies, regime-specific strategies, and historical trade-offs. In the first case, the agent retrieves knowledge about the coupling between signal control and bus scheduling under low-demand conditions, enabling coordinated optimization across both subsystems. In the second case, PSM recalls night-period traffic patterns and guides conservative signal adjustments suited to low-volume dynamics. In both cases, the agent improves stability and decision quality by reusing accumulated experience rather than relying solely on reactive reasoning, demonstrating continual self-improvement through procedural memory accumulation and adaptation.

\begin{figure}[t]
\centering
\begin{tcolorbox}[
  width=\columnwidth,
  center,
  enhanced,
  arc=2.2pt,
  boxrule=0.45pt,
  colframe=black!18,
  colback=white,
  left=4pt,
  right=4pt,
  top=4pt,
  bottom=4pt,
  boxsep=0pt,
  shadow={0.35mm}{-0.35mm}{0mm}{black!10},
]
\scriptsize
\setlength{\parskip}{0pt}
{\sffamily\scriptsize\bfseries\color{black!65}Case 1: Cross-subsystem dependency}\par\vspace{0.2em}
\begin{tcolorbox}[
  enhanced,
  frame hidden,
  borderline west={2.2pt}{0pt}{teal!55!black},
  colback=teal!6,
  arc=2pt,
  boxrule=0pt,
  left=5pt,
  right=4pt,
  top=3pt,
  bottom=3pt,
  before skip=0pt,
  after skip=3pt,
]
{\sffamily\scriptsize\bfseries\color{teal!45!black}Procedural memory}\par\vspace{0.1em}
{\scriptsize\itshape
\begin{itemize}[leftmargin=1.0em,itemsep=0.08em,topsep=0.05em,parsep=0pt]
\item Aggressive cycle time increases or headway reductions increase waiting time despite improving travel time or fuel consumption.
\item Signal control and bus scheduling are interdependent in low-demand periods; changes in one affect the other.
\end{itemize}}
\end{tcolorbox}
\begin{tcolorbox}[
  enhanced,
  frame hidden,
  borderline west={2.2pt}{0pt}{blue!45!black},
  colback=blue!4,
  arc=2pt,
  boxrule=0pt,
  left=5pt,
  right=4pt,
  top=3pt,
  bottom=3pt,
  before skip=0pt,
  after skip=3pt,
]
{\sffamily\scriptsize\bfseries\color{blue!40!black}Agent action}{\scriptsize\color{black!45}~(excerpt)}\par\vspace{0.15em}
{\scriptsize\ttfamily\upshape\raggedright\color{black!88}%
I'll start by planning my optimization strategy for the night period (03:30--04:00). Since \texttt{signal\_timing} affects \texttt{bus\_scheduling}, I should optimize \texttt{signal\_timing} first, then \texttt{bus\_scheduling}. I need to understand the current performance and constraints for both modules.}
\end{tcolorbox}
{\sffamily\scriptsize\bfseries\color{black!55}Results}\par\vspace{0.15em}
\begin{tcolorbox}[
  enhanced,
  colframe=black!12,
  colback=gray!4,
  arc=2pt,
  boxrule=0.35pt,
  left=4pt,
  right=4pt,
  top=3pt,
  bottom=3pt,
  before skip=0pt,
  after skip=3pt,
]
{\scriptsize\sffamily\bfseries\color{black!60}Signal control:}\par
{\scriptsize\textcolor{deepgreen}{$\checkmark$}~Travel time \textbf{449.65\,s} \textcolor{black!42}{\scriptsize (vs. previous best \ 452.95\,s, -3.30\,s)}}\par
{\scriptsize\textcolor{deepgreen}{$\checkmark$}~Waiting time \textbf{72.14\,s} \textcolor{black!42}{\scriptsize (vs. previous best \ 80.68\,s, -8.54\,s)}}\par\vspace{0.15em}
{\scriptsize\sffamily\bfseries\color{black!60}Bus scheduling:}\par
{\scriptsize\textcolor{deepgreen}{$\checkmark$}~Fuel consumption \textbf{112.64\,kg} \textcolor{black!42}{\scriptsize (vs. previous best \ 113.97\,kg, -1.33\,kg)}}\par
{\scriptsize\textcolor{deepgreen}{$\checkmark$}~Waiting time \textbf{434.19\,s} \textcolor{black!42}{\scriptsize (vs. previous best \ 533.54\,s, -99.35\,s)}}
\end{tcolorbox}

\vspace{0.2em}
\noindent\color{black!18}\rule{\linewidth}{0.35pt}
\vspace{0.2em}

{\sffamily\scriptsize\bfseries\color{black!65}Case 2: Night-period conservative optimization}\par\vspace{0.2em}
\begin{tcolorbox}[
  enhanced,
  frame hidden,
  borderline west={2.2pt}{0pt}{teal!55!black},
  colback=teal!6,
  arc=2pt,
  boxrule=0pt,
  left=5pt,
  right=4pt,
  top=3pt,
  bottom=3pt,
  before skip=0pt,
  after skip=3pt,
]
{\sffamily\scriptsize\bfseries\color{teal!45!black}Procedural memory}\par\vspace{0.1em}
{\scriptsize\itshape
\begin{itemize}[leftmargin=1.0em,itemsep=0.08em,topsep=0.05em,parsep=0pt]
\item Night period optimization requires minimal signal timing adjustments with fixed cycle times (60--90\,s) and equal distribution of green times.
\item Night period traffic patterns (low volume, minimal arrivals) require different optimization strategies than peak hours.
\end{itemize}}
\end{tcolorbox}
\begin{tcolorbox}[
  enhanced,
  frame hidden,
  borderline west={2.2pt}{0pt}{blue!45!black},
  colback=blue!4,
  arc=2pt,
  boxrule=0pt,
  left=5pt,
  right=4pt,
  top=3pt,
  bottom=3pt,
  before skip=0pt,
  after skip=3pt,
]
{\sffamily\scriptsize\bfseries\color{blue!40!black}Agent action}{\scriptsize\color{black!45}~(excerpt)}\par\vspace{0.15em}
{\scriptsize\ttfamily\upshape\raggedright\color{black!88}%
Night period optimization strategy: increase green time for high-demand phases, minimize low-demand phases at free-flow intersections, and reduce cycle times for night traffic.}
\end{tcolorbox}
{\sffamily\scriptsize\bfseries\color{black!55}Results}\par\vspace{0.15em}
\begin{tcolorbox}[
  enhanced,
  colframe=black!12,
  colback=gray!4,
  arc=2pt,
  boxrule=0.35pt,
  left=4pt,
  right=4pt,
  top=3pt,
  bottom=3pt,
  before skip=0pt,
  after skip=0pt,
]
{\scriptsize\sffamily\bfseries\color{black!60}Signal control}\par\vspace{0.1em}
{\scriptsize\textcolor{deepgreen}{$\checkmark$}~Travel time \textbf{466.48\,s} \textcolor{black!42}{\scriptsize (vs. previous best \ 478.18\,s, -11.70\,s)}}\par
{\scriptsize\textcolor{deepgreen}{$\checkmark$}~Waiting time \textbf{61.30\,s} \textcolor{black!42}{\scriptsize (vs. previous best \ 91.74\,s, -30.43\,s)}}
\end{tcolorbox}
\end{tcolorbox}
\caption{Case study of self-improvement.}
\Description{Styled card with three blocks: procedural memory with teal accent, agent plan in monospace with blue accent, and two side-by-side result panels for signal timing and bus scheduling with checkmarks.}
\label{fig:case_study_psm}
\vspace{-10pt}
\end{figure}

\section{Confidence Interval of Evaluation}\label{app:confidence-interval}

Table~\ref{tab:confidence_interval} reports the confidence intervals of TrafficClaw computed from three independent experimental runs with different random seeds on SUMO, covering both single-subsystem and multi-subsystem cooperative settings.

\begin{table}[t]
  \centering
  \caption{Confidence interval of TrafficClaw.}
  \label{tab:confidence_interval}
  \small
  \setlength{\tabcolsep}{3pt}
  \resizebox{\columnwidth}{!}{
  \begin{tabular}{l|c|c|c}
    \toprule
    \textbf{Metric} & \textbf{Signal Control} & \textbf{Bus--Signal} & \textbf{Highway--Ramp} \\
    \midrule
    Travel Time & 482.97 ($\pm 3.74$) & 459.87 ($\pm 5.09$) & 504.62 ($\pm 2.08$) \\
    Waiting Time & 133.84 ($\pm 0.99$) & 113.01 ($\pm 2.80$) & 413.38 ($\pm 6.26$) \\
    \bottomrule
  \end{tabular}}
  \vspace{-10pt}
\end{table}

\section{Experiment Configuration}\label{app:experiment-configuration}

We train TrafficClaw using VeRL~\cite{sheng2024hybridflow} and evaluate both open-source and proprietary LLMs through the SiliconFlow and OpenAI APIs, respectively. All evaluations use a fixed decoding temperature of 0.0 to ensure deterministic and reproducible results. The reference trajectories for multi-task cold-start are generated by the DeepSeek-V3.2. The decision horizon for all tasks is set to 30 minutes. TrafficClaw supports city-scale deployment on a single NVIDIA A800 GPU, requiring approximately 10.37s for each reasoning episode and 114.11s for each rollout.

\section{License}

Our code and the post-trained TrafficClaw model are released under the MIT License. For third-party models, Qwen3-Series \cite{yang2025qwen3} is licensed under Apache-2.0, DeepSeek-V3.2 \cite{liu2025deepseek} under MIT, and MiniMax-M2.5 \cite{minimax_m2_2025} and Kimi-K2.5 \cite{team2026kimi} under Modified MIT licenses; o4-mini \cite{openai_o3_o4_mini_2025} and Gemini-3.1-Pro \cite{deepmind_gemini} are proprietary API models used under the corresponding OpenAI and Google service terms. For third-party code and tools, VeRL \cite{sheng2024hybridflow} is Apache-2.0, AIDE \cite{jiang2025aide} is MIT, TrafficGPT \cite{zhang2024trafficgpt} is GPL-3.0, and SUMO \cite{alvarez_lopez_2026_18406080} is EPL-2.0 with GPL v2-or-later as a secondary license option; when a baseline repository \cite{wei2019colight,mnih2013playing,wang2022robust,wang2022shortening,xu2022novel,wu2020differential} does not declare an explicit license, we use it only for evaluation and do not redistribute its code. For data, NYC Mobility Survey \cite{nyc_mobility_survey}, and NYC TLC trip records \cite{nyc_tlc_trip_records} are public NYC Open Data resources governed by NYC terms of use, OpenStreetMap data \cite{osmplanet} is licensed under ODbL-1.0, and MTA schedule/open-data feeds \cite{mta_open_data} are used under the MTA data feed terms and conditions.
These existing artifacts are used consistently with their stated intended use: open-weight models are used as research baselines or training backbones under their model licenses, proprietary models are accessed only through their official APIs, third-party code and simulators are used for training or evaluation under their software licenses, and public transportation datasets are used for research simulation and evaluation under the corresponding data-provider terms.

\section{Data Anonymization}

The traffic demand data used in our experiments is sourced from publicly available NYC Open Data \cite{nyc_mobility_survey, nyc_tlc_trip_records}. The dataset is anonymized and does not contain any personally identifiable information regarding drivers or passengers.

\section{AI Assistant Usage}

We used AI assistants to polish the writing of this paper and to improve the visual presentation of figures. The figure-related assistance was limited to artistic styling and layout refinement; all underlying data, quantitative values, experimental results, and plots are obtained from the original outputs.

\section{Agent Prompt}\label{app:agent-prompt}

We provide the prompt used in TrafficClaw and example LLM outputs below.

\clearpage
\begin{figure*}[p]
\centering
\begin{tcolorbox}[
  width=0.98\textwidth,
  breakable,
  center,
  enhanced,
  arc=2.2pt,
  boxrule=0.45pt,
  colframe=black!18,
  colback=white,
  left=4pt,
  right=4pt,
  top=4pt,
  bottom=4pt,
  boxsep=0pt,
  shadow={0.35mm}{-0.35mm}{0mm}{black!10},
]
\scriptsize
\setlength{\parskip}{0.15em}
{\sffamily\scriptsize\bfseries\color{black!65}System prompt}\par\vspace{0.25em}
\begin{tcolorbox}[
  breakable,
  enhanced,
  frame hidden,
  borderline west={2.2pt}{0pt}{teal!55!black},
  colback=teal!6,
  arc=2pt,
  boxrule=0pt,
  left=5pt,
  right=4pt,
  top=3pt,
  bottom=3pt,
  before skip=0pt,
  after skip=4pt,
]
{\sffamily\scriptsize\bfseries\color{teal!45!black}\# Urban Transportation Joint Control Agent}\par\vspace{0.35em}
{\scriptsize\sffamily
\textbf{\#\# Your Role}\par
You are an expert urban transportation control agent optimizing multiple systems jointly.
Your goal is to coordinate multiple control modules to improve overall urban mobility.\par\vspace{0.35em}
\textbf{\#\# Global Knowledge}\par\vspace{0.2em}
\textbf{\#\#\# Enabled Modules}\par
\begin{itemize}[leftmargin=1.0em,itemsep=0.06em,topsep=0.05em,parsep=0pt]
\item \texttt{signal\_timing}
\item \texttt{bus\_scheduling}
\end{itemize}
(Only these modules can be optimized.)\par\vspace{0.25em}
\textbf{\#\#\# Cross-Module Dependencies}\par
\begin{itemize}[leftmargin=1.0em,itemsep=0.06em,topsep=0.05em,parsep=0pt]
\item \texttt{signal\_timing}: affects: \texttt{bus\_scheduling}
\item \texttt{bus\_scheduling}: affected\_by: \texttt{signal\_timing}
\end{itemize}\par\vspace{0.25em}
\textbf{\#\#\# Shared Optimization Principles}\par
\begin{itemize}[leftmargin=1.0em,itemsep=0.06em,topsep=0.05em,parsep=0pt]
\item Optimize upstream modules first (modules with no dependencies)
\item Downstream modules should consider upstream module decisions
\item Coordinate related modules for better overall performance
\end{itemize}\par\vspace{0.35em}
\textbf{\#\# Turn Limit}\par
You have a maximum of 20 dialogue turns.
Each action counts as one turn. Plan efficiently and use \texttt{FINISH} when complete.\par\vspace{0.35em}
\textbf{\#\# Available Actions}\par
\begin{enumerate}[leftmargin=1.2em,itemsep=0.08em,topsep=0.05em,parsep=0pt,label=\arabic*.]
\item \textbf{PLAN}: Think and plan your optimization strategy
\item \textbf{GET\_CONTROL\_API}: Query module-specific APIs, data, and domain knowledge\newline
   Available modules: \texttt{signal\_timing}, \texttt{bus\_scheduling}
\item \textbf{DATA\_ANALYSIS}: Analyze traffic data (use \texttt{save\_cache()} to store results)
\item \textbf{POLICY\_PLANNING}: Design control configurations (simulation runs automatically)
\item \textbf{DEBUG}: Fix code errors
\item \textbf{FINISH}: Complete optimization
\end{enumerate}\par\vspace{0.35em}
\textbf{\#\# Previous Optimization Experience (Memory)}\par
\begin{enumerate}[leftmargin=1.2em,itemsep=0.08em,topsep=0.05em,parsep=0pt]
\item Conservative optimization (5--10\% changes) is more effective than aggressive changes for both signal timing and bus scheduling.
\item Targeted adjustments (top 50 intersections, top 30 routes) work better than broad changes, focusing on queue $>30$ or waiting time $>80$ thresholds.
\item Signal timing optimization should precede bus scheduling optimization due to dependency relationships.
\item Signal timing improvements often come with waiting time trade-offs but can maintain throughput.
\item Bus scheduling improvements can increase fuel consumption but reduce passenger waiting time significantly.
\item Accept 1--2\% reward decrease if specific metrics (like passenger waiting time) improve significantly.
\item Use 85--90\% headway reductions for bus scheduling adjustments based on waiting time thresholds ($>$\,150\,s).
\item Midday period (12:30--13:00) shows consistent traffic patterns suitable for targeted optimization.
\item Focus on intersections with queue $>30$ or waiting time $>80$ for signal timing adjustments.
\item Network has complex intersections with multiple phase types requiring careful signal timing adjustments. 
\end{enumerate}\par\vspace{0.35em}
\textbf{\#\# Important Notes}\par
\begin{itemize}[leftmargin=1.0em,itemsep=0.06em,topsep=0.05em,parsep=0pt]
\item Use \texttt{GET\_CONTROL\_API} to query module APIs and domain knowledge before optimization
\item Only enabled modules can be optimized
\item Consider module dependencies when planning optimization order
\item Simulation is automatically executed after \texttt{POLICY\_PLANNING}
\end{itemize}\par\vspace{0.35em}
\textbf{\#\# Required Formats (Critical)}\par
\begin{itemize}[leftmargin=1.0em,itemsep=0.06em,topsep=0.05em,parsep=0pt]
\item Always start with: \texttt{ACTION: <ACTION\_NAME>}
\item \texttt{GET\_CONTROL\_API} format, \texttt{DATA\_ANALYSIS} format, \texttt{POLICY\_PLANNING} format, \texttt{DEBUG} (policy) format (see listing below)
\item One action per turn. Do not combine actions in a single response.
\end{itemize}
\begin{lstlisting}[
  language=Python,
  breaklines=true,
  basicstyle=\scriptsize\ttfamily,
  columns=fullflexible,
  keepspaces=true,
  showstringspaces=false,
  frame=single,
  rulecolor=\color{black!15},
  linewidth=\dimexpr\linewidth-1.5em\relax,
]
# Required action format examples (verbatim from the system prompt).
# GET_CONTROL_API format:
ACTION: GET_CONTROL_API
Module: signal_timing, bus_scheduling  # comma-separated allowed

# DATA_ANALYSIS format:
ACTION: DATA_ANALYSIS
# Use save_cache(...) to store analysis results

# POLICY_PLANNING format:
ACTION: POLICY_PLANNING
Control Modules: signal_timing, bus_scheduling
# Define a variable named `config` and do NOT use return
config = {...}

# DEBUG (policy) format:
ACTION: DEBUG
Debug Type: POLICY_PLANNING
Control Modules: signal_timing, bus_scheduling
# Fixed code
\end{lstlisting}\par\vspace{0.35em}
Begin with \texttt{PLAN} to outline your optimization strategy, then use \texttt{GET\_CONTROL\_API} to query module APIs.}
\end{tcolorbox}
\end{tcolorbox}
\end{figure*}

\clearpage
\begin{figure*}[p]
\centering
\begin{tcolorbox}[
  width=0.98\textwidth,
  breakable,
  center,
  enhanced,
  arc=2.2pt,
  boxrule=0.45pt,
  colframe=black!18,
  colback=white,
  left=4pt,
  right=4pt,
  top=4pt,
  bottom=4pt,
  boxsep=0pt,
  shadow={0.35mm}{-0.35mm}{0mm}{black!10},
]
\scriptsize
{\sffamily\scriptsize\bfseries\color{black!65}Agent Response: Executable Spatiotemporal Analysis}\par\vspace{0.12em}
\begin{tcolorbox}[
  breakable,
  enhanced,
  frame hidden,
  borderline west={2.2pt}{0pt}{blue!45!black},
  colback=blue!4,
  arc=2pt,
  boxrule=0pt,
  left=5pt,
  right=4pt,
  top=3pt,
  bottom=3pt,
  before skip=0pt,
  after skip=2pt,
]
{\sffamily\scriptsize\bfseries\color{blue!40!black}Agent response}{\scriptsize\color{black!45}~(abbreviated turn)}\par\vspace{0.1em}
{\scriptsize\ttfamily\upshape\raggedright\color{black!88}%
ACTION: DATA\_ANALYSIS}
\begin{lstlisting}[
  language=Python,
  breaklines=true,
  basicstyle=\scriptsize\ttfamily,
  columns=fullflexible,
  keepspaces=true,
  showstringspaces=false,
  frame=single,
  rulecolor=\color{black!15},
  linewidth=\dimexpr\linewidth-1.5em\relax,
]
# DATA_ANALYSIS (abbreviated): joint signal_timing + bus_scheduling analytics
print("=== Signal Timing Analysis ===")
signal_data = read_lane_traffic_states()
if 'error' not in signal_data:
    snapshots = signal_data.get('snapshots', [])
    if snapshots:
        lane_states = snapshots[-1].get('lane_states', {})
        intersection_queues, phase_demands = {}, {}
        for lane_id, lane_data in lane_states.items():
            q = lane_data.get('queue_length', 0)
            end_inter = lane_data.get('end_intersection')
            loc_dir = lane_data.get('loc_dir', '')
            if end_inter and end_inter != 'None':
                intersection_queues.setdefault(end_inter, []).append(q)
            if loc_dir and end_inter and end_inter != 'None':
                key = (end_inter, loc_dir)
                phase_demands[key] = phase_demands.get(key, 0) + q

        intersection_avg_queue = {
            i: sum(qs) / len(qs) for i, qs in intersection_queues.items() if qs
        }
        busy = [i for i, avg in intersection_avg_queue.items() if avg >= 5]
        free = [i for i, avg in intersection_avg_queue.items() if avg < 5]
        save_cache({
            'busy_intersections': busy,
            'free_flow_intersections': free,
            'phase_demands': phase_demands,
            'intersection_avg_queue': intersection_avg_queue,
        })

print("\n=== Bus Scheduling Analysis ===")
bus_data = read_bus_states()
if 'error' not in bus_data:
    snapshots = bus_data.get('snapshots', [])
    if snapshots:
        bus_lines = snapshots[-1].get('bus', {}).get('lines', {})
        station_waiting_analysis, route_headway_analysis = {}, {}
        save_cache({'route_headway_analysis': route_headway_analysis})  # ... stations / routes omitted

print("\nData analysis complete. Results saved to cache.")
\end{lstlisting}
\end{tcolorbox}
\end{tcolorbox}

\vspace{0.1em}

\begin{tcolorbox}[
  width=0.98\textwidth,
  breakable,
  center,
  enhanced,
  arc=2.2pt,
  boxrule=0.45pt,
  colframe=black!18,
  colback=white,
  left=4pt,
  right=4pt,
  top=4pt,
  bottom=4pt,
  boxsep=0pt,
  shadow={0.35mm}{-0.35mm}{0mm}{black!10},
]
\scriptsize
{\sffamily\scriptsize\bfseries\color{black!65}Agent Response: Feedback-Driven decision-making}\par\vspace{0.12em}
\begin{tcolorbox}[
  breakable,
  enhanced,
  frame hidden,
  borderline west={2.2pt}{0pt}{blue!45!black},
  colback=blue!4,
  arc=2pt,
  boxrule=0pt,
  left=5pt,
  right=4pt,
  top=3pt,
  bottom=3pt,
  before skip=0pt,
  after skip=2pt,
]
{\sffamily\scriptsize\bfseries\color{blue!40!black}Agent response}{\scriptsize\color{black!45}~(abbreviated turn)}\par\vspace{0.1em}
{\scriptsize\ttfamily\upshape\raggedright\color{black!88}%
ACTION: POLICY\_PLANNING\par
Control Modules: \texttt{signal\_timing}, \texttt{bus\_scheduling}}
\begin{lstlisting}[
  language=Python,
  breaklines=true,
  basicstyle=\scriptsize\ttfamily,
  columns=fullflexible,
  keepspaces=true,
  showstringspaces=false,
  frame=single,
  rulecolor=\color{black!15},
  linewidth=\dimexpr\linewidth-1.5em\relax,
]
# POLICY_PLANNING (abbreviated): write executable configs for enabled modules
import copy
signal_config = current_signal_config.copy()
phase_demands_cache = load_cache('phase_demands')
for inter_id, phases in signal_config.items():
    current_phases = current_signal_config[inter_id]
    phase_demand_map = {}
    if phase_demands_cache and f'phase_demand_{inter_id}' in phase_demands_cache:
        phase_demand_map = phase_demands_cache[f'phase_demand_{inter_id}']
    total_demand = sum(phase_demand_map.values())
    if total_demand > 0:
        new_durations = {}
        for phase_name, dur in current_phases.items():
            d = phase_demand_map.get(phase_name, 0)
            new_durations[phase_name] = max(10, int(dur * (d / total_demand))) if d > 0 else 10
        signal_config[inter_id] = new_durations  # ... cycle-time scaling omitted
    else:
        signal_config[inter_id] = {p: max(10, current_phases[p]) for p in current_phases}

bus_config = {}
for route_id, config in current_bus_schedule.items():
    bus_config[route_id] = {"timetable": []}
    for period in config['timetable']:
        new_period = {
            "time_range": period["time_range"],
            "headway": period["headway"],
            "schedule": copy.deepcopy(period["schedule"]),
        }
        bus_config[route_id]["timetable"].append(new_period)  # ... dwell/headway edits omitted
config = {"signal_timing": signal_config, "bus_scheduling": bus_config}
\end{lstlisting}
\end{tcolorbox}
\end{tcolorbox}
\end{figure*}

\clearpage
\begin{figure*}[p]
\centering
\begin{tcolorbox}[
  width=0.98\textwidth,
  breakable,
  center,
  enhanced,
  arc=2.2pt,
  boxrule=0.45pt,
  colframe=black!18,
  colback=white,
  left=4pt,
  right=4pt,
  top=4pt,
  bottom=4pt,
  boxsep=0pt,
  shadow={0.35mm}{-0.35mm}{0mm}{black!10},
]
\scriptsize
\setlength{\parskip}{0.1em}
{\sffamily\scriptsize\bfseries\color{black!65}Procedural Spatiotemporal Memory Update Prompt}\par\vspace{0.15em}
\begin{tcolorbox}[
  breakable,
  enhanced,
  frame hidden,
  borderline west={2.2pt}{0pt}{teal!55!black},
  colback=teal!6,
  arc=2pt,
  boxrule=0pt,
  left=5pt,
  right=4pt,
  top=3pt,
  bottom=3pt,
  before skip=0pt,
  after skip=2pt,
]
{\scriptsize\sffamily
[You have just completed an optimization session. Above is the complete conversation history including all your actions, code executions, simulation results, and feedback.\par\vspace{0.15em}
\textbf{Session Summary} (for quick reference, but review full conversation above):\par
\begin{itemize}[leftmargin=1.0em,itemsep=0.04em,topsep=0.03em,parsep=0pt]
\item Total turns: 13
\item Actions taken: PLAN(4), GET\_CONTROL\_API(1), DATA\_ANALYSIS(3), POLICY\_PLANNING(3), DEBUG(1), FINISH(1)
\end{itemize}\par\vspace{0.12em}
\textbf{Best Module-level Results:}\par
\begin{itemize}[leftmargin=1.0em,itemsep=0.04em,topsep=0.03em,parsep=0pt]
\item Signal timing (best turn 5): reward 218{,}246.00; avg queue 8.43; avg waiting 59.87\,s; throughput 32.00.
\item Bus scheduling (best turn 5): reward 218{,}246.00; avg passenger waiting time 179.09\,s; fuel 102{,}208.54\,g; on-time rate 0.10; 1 module improved vs.\ baseline (bus).
\end{itemize}\par\vspace{0.15em}
\textbf{Your Reflection Task (Multi-turn, using DATA\_ANALYSIS):}\par
\begin{enumerate}[leftmargin=1.2em,itemsep=0.05em,topsep=0.03em,parsep=0pt]
\item Enter a short reflection phase (up to \textbf{5} \texttt{DATA\_ANALYSIS} turns).
\item Re-use cached sandbox data via \texttt{save\_cache()}/\texttt{load\_cache()}/\texttt{list\_cache()}; do \emph{not} run \texttt{POLICY\_PLANNING}.
\item Final response: \texttt{ACTION: REFLECTION\_FINISH} with \textbf{only} an updated memory list (JSON array of strings in a \texttt{json} block).
\end{enumerate}\par\vspace{0.12em}
\textbf{Final Output Format (Last Reflection Response with REFLECTION\_FINISH):}\par
\texttt{ACTION: REFLECTION\_FINISH}\par\vspace{0.08em}
\begin{lstlisting}[
  basicstyle=\scriptsize\ttfamily,
  breaklines=true,
  columns=fullflexible,
  frame=single,
  rulecolor=\color{black!15},
  linewidth=\dimexpr\linewidth-1.5em\relax,
]
[
    "INSIGHT 1",
    "INSIGHT 2",
    "...",
    "INSIGHT N"
]
\end{lstlisting}\par\vspace{0.12em}
\textbf{Important:}\par
\begin{itemize}[leftmargin=1.0em,itemsep=0.04em,topsep=0.03em,parsep=0pt]
\item Return a valid JSON array (max.\ 10 items): keep/refine prior memory, add new insights, drop unwanted items by omission.
\item One sentence per item; robust, transferable strategies; no extra commentary outside the JSON array.
\end{itemize}\par}
\end{tcolorbox}
\vspace{0.18em}
{\sffamily\scriptsize\bfseries\color{black!65}Example reflection outputs}\par\vspace{0.08em}
\begin{tcolorbox}[
  breakable,
  enhanced,
  frame hidden,
  borderline west={2.2pt}{0pt}{blue!45!black},
  colback=blue!4,
  arc=2pt,
  boxrule=0pt,
  left=5pt,
  right=4pt,
  top=2pt,
  bottom=2pt,
  before skip=0pt,
  after skip=2pt,
]
{\sffamily\scriptsize\bfseries\color{blue!40!black}Agent Response: Procedural Spatiotemporal Memory Summary}\par\vspace{0.08em}
{\scriptsize\ttfamily\upshape\raggedright\color{black!88}%
ACTION: DATA\_ANALYSIS}
\begin{lstlisting}[
  language=Python,
  breaklines=true,
  basicstyle=\scriptsize\ttfamily,
  columns=fullflexible,
  keepspaces=true,
  showstringspaces=false,
  frame=single,
  rulecolor=\color{black!15},
  linewidth=\dimexpr\linewidth-1.5em\relax,
]
# Reflection DATA_ANALYSIS (abbreviated): post-session pattern mining
print("=== REFLECTION: NETWORK PATTERNS ===")
try:
    cached_data = load_cache()
except Exception:
    cached_data = {}
intersection_types = {}
traffic_data = read_traffic_states(start_time=45001 - 3600, end_time=45001)
if 'error' not in traffic_data:
    snapshots = traffic_data.get('snapshots', [])
    time_periods = {}
    for snapshot in snapshots:
        hour = int(snapshot.get('time', 0) // 3600)
        bucket = time_periods.setdefault(hour, {"vehicles": 0, "waiting": 0, "n": 0})
        vehicles = snapshot.get('vehicles', [])
        bucket["vehicles"] += len(vehicles)
        bucket["waiting"] += sum(1 for v in vehicles if v.get('state') == 'waiting')
        bucket["n"] += 1
    # ... omitted: hourly printouts and intersection-type statistics
save_cache({
    "traffic_patterns": time_periods if 'time_periods' in locals() else {},
    "intersection_types": intersection_types if 'intersection_types' in locals() else {},
})
\end{lstlisting}
\end{tcolorbox}
\vspace{0.15em}
\begin{tcolorbox}[
  breakable,
  enhanced,
  frame hidden,
  borderline west={2.2pt}{0pt}{blue!45!black},
  colback=blue!4,
  arc=2pt,
  boxrule=0pt,
  left=5pt,
  right=4pt,
  top=2pt,
  bottom=2pt,
  before skip=0pt,
  after skip=2pt,
]
{\sffamily\scriptsize\bfseries\color{blue!40!black}Agent Response: Procedural Spatiotemporal Memory Update}\par\vspace{0.08em}
{\scriptsize\ttfamily\upshape\raggedright\color{black!88}%
ACTION: REFLECTION\_FINISH}
\begin{lstlisting}[
  basicstyle=\scriptsize\ttfamily,
  breaklines=true,
  columns=fullflexible,
  frame=single,
  rulecolor=\color{black!15},
  linewidth=\dimexpr\linewidth-1.5em\relax,
]
[
    "Bus scheduling optimization during morning rush hour (07:30-08:00) should focus on aggressive headway reduction (25-30%) to reduce passenger waiting time, even if it increases fuel consumption.",
    "Signal timing optimization requires demand-based analysis rather than simple cycle time increases; focus on intersections with clear demand patterns (>10 queue length) for effective improvements.",
    "The network has short signal cycles (average 29.5 seconds) with most phases at minimum 10 seconds, indicating conservative signal timing by default.",
    "During rush hours, bus headways should be reduced to 60-180 seconds to minimize waiting time, with dwell time increases at high-demand stations.",
    "There is a trade-off between waiting time reduction and fuel consumption; prioritize waiting time reduction as the primary objective for public transit service quality.",
    "Bus scheduling optimization should target specific time ranges (e.g., 07:30-08:00) with time-based headway adjustments rather than uniform changes across all periods.",
    "Signal timing optimization should consider the optimization window's time period (morning rush) and adjust strategies accordingly, not just apply general increases.",
    "The network's bus routes have default headways of 600 seconds (10 minutes), indicating potential for significant improvement during peak demand periods.",
    "When optimizing multiple modules jointly, prioritize upstream modules (signal timing) first, but in this case, bus scheduling improvements had a more immediate impact.",
    "Future optimizations should use historical traffic data analysis to identify specific intersections and routes needing adjustment rather than blanket changes."
]
\end{lstlisting}
\end{tcolorbox}
\end{tcolorbox}
\end{figure*}
\clearpage
\begin{figure*}[p]
\centering
\begin{tcolorbox}[
  width=0.98\textwidth,
  breakable,
  center,
  enhanced,
  arc=2.2pt,
  boxrule=0.45pt,
  colframe=black!18,
  colback=white,
  left=4pt,
  right=4pt,
  top=4pt,
  bottom=4pt,
  boxsep=0pt,
  shadow={0.35mm}{-0.35mm}{0mm}{black!10},
]
\scriptsize
\setlength{\parskip}{0.1em}
{\sffamily\scriptsize\bfseries\color{black!65}LLM-as-Judge Prompt: Policy Consistency and System-Level Traffic Efficiency Alignment}\par\vspace{0.1em}
\begin{tcolorbox}[
  breakable,
  enhanced,
  frame hidden,
  borderline west={2.2pt}{0pt}{orange!60!black},
  colback=orange!4,
  arc=2pt,
  boxrule=0pt,
  left=5pt,
  right=4pt,
  top=3pt,
  bottom=3pt,
  before skip=0pt,
  after skip=2pt,
]
{\scriptsize\sffamily
\texttt{You are a professional traffic optimization expert. Please evaluate the following conversation between an LLM Agent and a traffic simulation environment. Score from 0-10 points.}\par\vspace{0.06em}
\textbf{Dimension 1: Multi-module Coordination Quality (0--5 points):}\par
\begin{itemize}[leftmargin=1.0em,itemsep=0.04em,topsep=0.03em,parsep=0pt]
\item This environment has the following control modules: \texttt{\{module\_names\_str\}}.
\item Whether these modules are reasonably coordinated with each other.
\item Whether the interactions and dependencies between modules are considered.
\item Whether conflicts between modules are identified and avoided.
\end{itemize}\par\vspace{0.12em}
\textbf{Dimension 2: Modeling Effectiveness (0--5 points):}\par
\begin{itemize}[leftmargin=1.0em,itemsep=0.04em,topsep=0.03em,parsep=0pt]
\item Whether the policy code correctly implements the optimization approach for \texttt{\{module\_names\_str\}}.
\item Whether appropriate algorithms and parameters are used.
\item Whether the data provided by the environment is fully utilized.
\item Whether the agent iteratively improves the policy based on simulation feedback.
\end{itemize}\par\vspace{0.12em}
\texttt{Please read the conversation carefully and provide your evaluation in the following STRICT format:}\par
\begin{lstlisting}[
  basicstyle=\scriptsize\ttfamily,
  breaklines=true,
  columns=fullflexible,
  frame=single,
  rulecolor=\color{black!15},
  linewidth=\dimexpr\linewidth-1.5em\relax,
]
Score: [0-10 integer]
Brief Comment: [Your brief comment in 1-2 sentences]
\end{lstlisting}\par\vspace{0.06em}
\texttt{Example output:}\par
\begin{lstlisting}[
  basicstyle=\scriptsize\ttfamily,
  breaklines=true,
  columns=fullflexible,
  frame=single,
  rulecolor=\color{black!15},
  linewidth=\dimexpr\linewidth-1.5em\relax,
]
Score: 7
Brief Comment: The agent demonstrates good optimization strategy with reasonable parameter tuning, but could better utilize the simulation feedback for iterative improvement.
\end{lstlisting}\par\vspace{0.06em}
\texttt{Conversation:}\par
\texttt{\{conversation\_text\}}\par\vspace{0.06em}
\texttt{Your Evaluation:}\par}
\end{tcolorbox}
\end{tcolorbox}
\end{figure*}

\end{document}